\pdfoutput=1
\documentclass[letterpaper]{article} 
\usepackage{aaai24}  
\usepackage{times}  
\usepackage{helvet}  
\usepackage{courier}  
\usepackage[hyphens]{url}  
\usepackage{graphicx} 
\usepackage{natbib}  
\usepackage{caption} 
\usepackage[linesnumbered,ruled,vlined]{algorithm2e}
\usepackage{booktabs}
\usepackage{pifont}
\usepackage{enumitem}
\newcommand{\cmark}{\color{green}{\ding{51}}}%
\newcommand{\xmark}{\color{red}{\ding{55}}}%
\newcommand{\vmark}{\ding{226}\hspace{1em}}
\usepackage[table]{xcolor}
\usepackage{array}

\usepackage{amssymb}
\usepackage{amsmath,amsfonts}
\usepackage{amsopn}
\usepackage{bm} 
\usepackage{multirow}
\usepackage{tabularx}
\usepackage{tikz}
\usepackage{xspace}

\newcommand{\PFLBed}{\method{PFLBed}\xspace}
\newcommand{\PerBasis}{\method{PerBasis}\xspace}
\newcommand{\FedBasis}{\method{FedBasis}\xspace}

\newcommand{\FedAvg}{\method{FedAvg}\xspace}

\newcommand{\pFedMe}{\method{pFedMe}\xspace}
\newcommand{\PerFedAvg}{\method{Per-FedAvg}\xspace}
\newcommand{\PerFedAvgFT}{\method{Per-FedAvg+FT}\xspace}

\newcommand{\FedBN}{\method{FedBN}\xspace}
\newcommand{\pFedHN}{\method{pFedHN}\xspace}
\newcommand{\FedRep}{\method{FedRep}\xspace}

\newcommand{\kNNPer}{\method{kNN-Per}\xspace}
\newcommand{\FedAvgLP}{\method{FedAvg+LP}\xspace}
\newcommand{\FedAvgFT}{\method{FedAvg+FT}\xspace}
\newcommand{\pFedHNLP}{\method{pFedHN+LP}\xspace}
\newcommand{\pFedHNFT}{\method{pFedHN+FT}\xspace}
\newcommand{\FedRepLP}{\method{FedRep+LP}\xspace}
\newcommand{\FedRepFT}{\method{FedRep+FT}\xspace}
\newcommand{\FedBNLP}{\method{FedBN+LP}\xspace}
\newcommand{\FedBNFT}{\method{FedBN+FT}\xspace}
\newcommand{\kNNPerFT}{\method{kNN-Per+FT}\xspace}

\newcommand{\vct}[1]{\boldsymbol{#1}} 
\newcommand{\mat}[1]{\boldsymbol{#1}} 

\newcommand{\field}[1]{\mathbb{#1}}
\newcommand{\R}{\field{R}} 

\newcommand{\ProbOpr}[1]{\mathbb{#1}}

\newcommand{\expect}[2]{%
\ifthenelse{\equal{#2}{}}{\ProbOpr{E}_{#1}}
{\ifthenelse{\equal{#1}{}}{\ProbOpr{E}\left[#2\right]}{\ProbOpr{E}_{#1}\left[#2\right]}}} 

\DeclareMathOperator{\argmin}{arg\,min}

\newcommand{\vtheta}{\vct{\theta}}

\newcommand{\vx}{{\vct{x}}}

\newcommand{\vv}{\vct{v}}

\newcommand{\vphi}{\vct{\phi}}
\newcommand{\vpsi}{\vct{\psi}}
\newcommand{\valpha}{\vct{\alpha}}

\newcommand{\mOmega}{\mat{\Omega}}

\newcommand{\sD}{\mathcal{D}}

\newcommand{\sL}{\mathcal{L}}
\newcommand{\sA}{\mathcal{A}}
\newcommand{\sR}{\mathcal{R}}
\newcommand{\sV}{\mathcal{V}}
\newcommand{\sP}{\mathcal{P}}
\newcommand{\eat}[1]{}
\newcommand{\method}[1]{\textsc{#1}}

\newcommand{\HYC}[1]{{\color{red} HY: #1}\xspace}

\newcommand{\ie}{i.e.\xspace}
\newcommand{\eg}{e.g.\xspace}

\usepackage[capitalize]{cleveref}
\crefname{section}{Sec.}{Secs.}
\Crefname{section}{Section}{Sections}
\Crefname{table}{Table}{Tables}
\crefname{table}{Tab.}{Tabs.}

\title{Federated Learning of Shareable Bases for\\ Personalization-Friendly Image Classification}
\author{Hong-You Chen$^1$\equalcontrib\thanks{Partial work done as a research intern at Google Research.}~~~~~
Jike Zhong$^1$\equalcontrib~~~~~ 
Mingda Zhang$^2$~~~~~
Xuhui Jia$^2$~~~~~\\
Hang Qi$^2$~~~~~
Boqing Gong$^2$~~~~~
Wei-Lun Chao$^1$~~~~~
Li Zhang$^2$~~~~~\\
}

\affiliations{
$^1$The Ohio State University~~~ $^2$Google Research
}

\begin{document}

\maketitle
\begin{abstract}
Personalized federated learning (PFL) aims to harness the collective wisdom of clients' data while building personalized models tailored to individual clients' data distributions. Existing works offer personalization primarily to clients who participate in the FL process, making it hard to encompass new clients who were absent or newly show up. In this paper, 
we propose \FedBasis, a novel PFL framework to tackle such a deficiency. \FedBasis learns a set of few shareable ``basis'' models, which can be linearly combined to form personalized models for clients. Specifically for a new client, only a small set of combination coefficients, not the model weights, needs to be learned. This notion makes \FedBasis more parameter-efficient, robust, and accurate than competitive PFL baselines, especially in the low data regime, without increasing the inference cost. To demonstrate the effectiveness and applicability of \FedBasis, we also present a more practical PFL testbed for image classification, featuring larger data discrepancies across clients in both the image and label spaces as well as more faithful training and test splits.
\end{abstract}

\section{Introduction}
\label{s-intro}

Recent years have witnessed a gradual shift in machine learning towards taking users' aspects into account. Building personalized models (\eg, image classifiers) tailored to users' data, preferences, and characteristics has been shown to improve user experience greatly~\cite {rudovic2018personalized}.

To achieve so, however, may sacrifice data privacy and ownership during the collection of training data, as highlighted in~\cite{jordan2015machine,papernot2016towards}. Personalized federated learning (PFL) is a promising machine learning paradigm that can fulfill the demands of both worlds~\cite{tan2022towards}. On the one hand, it strictly follows the setup of federated learning (FL): training models collaboratively with multiple users (\ie, clients) while keeping their data decentralized~\cite{kairouz2019advances}. On the other hand, it personalizes models for clients that feature better accuracy in their respective data distributions.

Despite making promising progress, existing works of PFL mostly limit their personalization capability to clients who participate in the FL process. For example, mainstream methods based on multi-task learning~\cite{li2020federated,smith2017federated} jointly train models for clients to prevent over-fitting.
For a new client who was not involved in the previous FL process, there is no clear principle to construct a personalized model except for conducting another run of the FL process. This deficiency greatly limits the applicability of PFL in practice, especially for a personalization service provider: new clients may show up at any time, and it is extremely inefficient to rerun the FL process every time.

At first glance, one may resolve this problem by first training a global model with participating clients (\eg, via \FedAvg~\cite{mcmahan2017communication}) and then fine-tuning it for each new client. However, it has two noticeable drawbacks. First, fine-tuning an over-parameterized neural network with limited data, which is often the case for new clients, is known to be sensitive to hyperparameters and prone to over-fitting~\cite{li2021ditto,pillutla2022federated,wu2022motley,fallah2020personalized}.
Second, fine-tuning a single global model prevents us from leveraging the diversity and relationships of the participating clients to facilitate personalization for new clients.

In this paper, we, therefore, strive to tackle a novel PFL problem: \emph{how to learn and leverage knowledge from participating clients in the past to facilitate personalization for new clients, especially in the low data regime?}

\begin{figure}[t]
    \centering
    \includegraphics[width=\columnwidth]{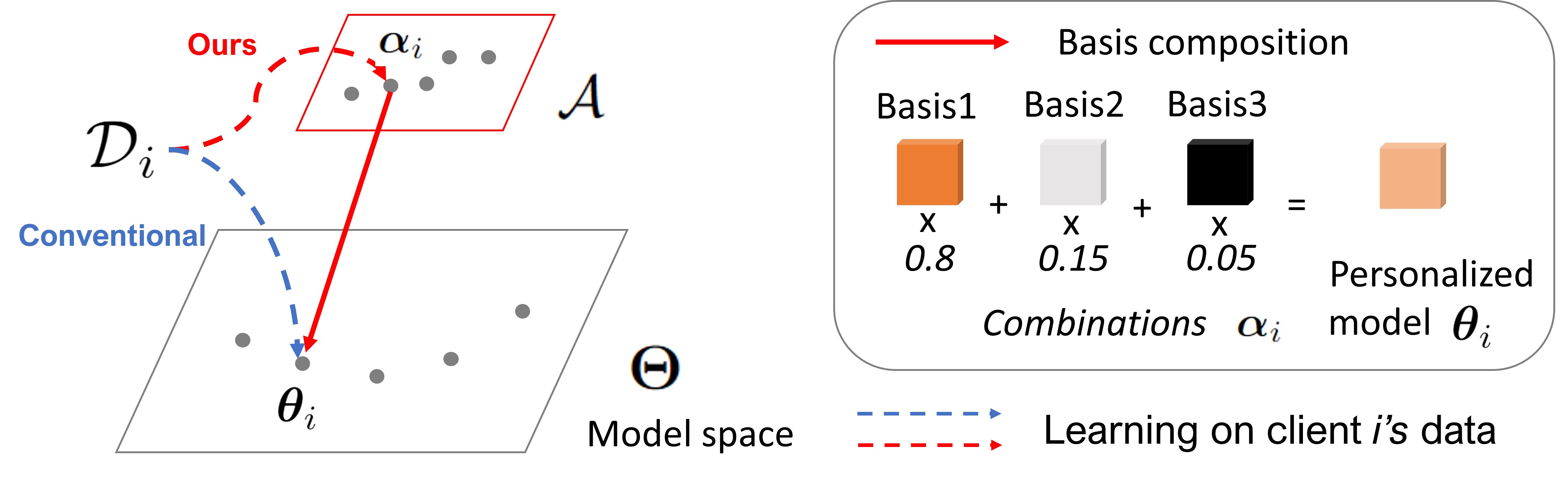}
    \vskip -8pt
    \caption{\small In conventional PFL methods, \emph{each client} learns a \emph{high-dimensional model} (blue) on its dataset $\sD_i$. In \textbf{our \FedBasis}, we learn a few shareable basis models of the same neural network architecture on the participating clients. After the basis models are trained, a \emph{new client} only needs to learn a \emph{low-dimensional vector} (red) as the coefficients to combine them into a personalized network, making \textbf{\FedBasis} more data-efficient and robust for supporting new clients, \ie, more ``personalization-friendly."} 
    \label{fig:conceptual}
    \vskip -15pt
\end{figure}

We make a mild assumption: the data distribution of a new client is covered by the aggregated data distribution of all the participating clients. More precisely, treating each client as a ``point'' in the ``client space'', we assume that a new client is located on or near the subspace spanned by the participating clients. Under this assumption, we propose a novel ``personalization-friendly'' PFL framework called \FedBasis, which goes beyond learning personalized models for participating clients (\ie, point estimates) toward learning the underlying subspace of clients' models so that we can rapidly and robustly construct personalized models for new clients. Concretely, \FedBasis learns \emph{a few shareable ``basis'' models} of the same architecture to capture the subspace spanned by clients, inspired by~\cite{changpinyo2016synthesized,evgeniou2007multi}. With these basis models, we can synthesize a client model by a linear (more strictly, convex) combination estimated on the fly. \cref{fig:conceptual} gives an illustration. 

Despite its conceptual simplicity, \FedBasis has several notable advantages for PFL. First, \FedBasis is reminiscent of dictionary learning~\cite{mairal2009online} in the neural network's parameter space. That is, \FedBasis reduces the overall learnable parameters by summarizing the participating clients' models into the basis models, effectively reducing the sample complexity in federated learning.
Second, when a new client arrives, \FedBasis learns the combination coefficients, not the model weights, to construct a personalized model, making it more robust to the low data regime. Importantly, \FedBasis combines the parameters of the basis models, not their predictions (sharply different from the mixture of experts~\cite{reisser2021federated}). As such, the inference cost remains almost the same as a single neural network model and does not scale with the number of bases.

To ensure that the basis models learn diverse knowledge in a federated setting to support a variety of client distributions\footnote{In an FL setting which involves local model training with each client's data alone, we found it challenging to learn diverse basis models --- the bases easily collapse into non-specialized models.}, we propose a \emph{coordinate descent} style model update. During each round of local training, we first update the combination coefficients alone, freeze and sharpen them (so most elements are near zero), and perform SGD solely on the basis models for multiple epochs.
The sharpening operation limits the bases each client can use and, in turn, forces different bases to learn
from different subsets of clients, leading to basis models that capture diverse specialized knowledge.

To support more realistic and faithful PFL evaluation, as a side contribution, we construct a new set of benchmark datasets, \PFLBed. The motivations for \PFLBed are two-fold. First, \PFLBed is carefully designed to minimize the mismatch between the training and testing distributions for each client. To our surprise, we found such a mismatch huge in existing datasets \cite{caldas2018leaf,li2021ditto}, which may mislead the progress of PFL. Second, we consider more challenging non-IID conditions across clients, capturing variations in both the data and label distributions, in contrast to many existing datasets that focus on merely one of them \cite{chen2022bridging,sun2021partialfed}. 

We validate \FedBasis on both the standard PFL datasets and \PFLBed for constructing personalized models for new clients. Compared to mainstream PFL approaches, \FedBasis achieves more robust performance across various PFL settings, demonstrating its superiority for personalization.

\section{Related Work}
\label{s_related}
Conventional PFL is quite well-studied for clients who participate in the federated learning process. Differently, we focus on a less explored problem that aims to construct personalized models for individual new clients, which was first raised briefly in~\cite{shamsian2021personalized,collins2021exploiting}.

\noindent \textbf{Conventional PFL.} 
Many earlier works formulate personalization with multiple clients as
\emph{multi-task learning (MTL)} and focus on regularizer designs while each client learns its own model~\cite{smith2017federated,zhang2021parameterized,li2021ditto,dinh2020personalized}. \emph{Mixture of models}~\cite{zec2020federated,marfoq2021federated,luo2022adapt,ruan2022fedsoft} assumes the clients' data are from a mixture of distributions. It then learns a global model and a set of local models and takes a mixture of them in their outputs (not the model weights like ours) to perform personalized predictions. 
\emph{Clustered FL}~\cite{ghosh2020efficient} relies on a rather strong assumption that the clients are grouped into a few clusters and share one model per cluster. 

\noindent \textbf{General / generalized representations.} More recent approaches rely on a universal feature extractor. Each client only personalizes an output head~\cite{collins2021exploiting,chen2022bridging}, a Gaussian process tree classifier~\cite{achituve2021personalized}, or a $k$-NN classifier~\cite{marfoq2022personalized}. Such an approach is simple and strong but likely sub-optimal when the features are required to be personalized. We agree on the concept of learning powerful representation but relax the single-model constraint by maintaining multiple shareable basis models. Another related topic \emph{federated domain generalization}~\cite{nguyen2022fedsr,zhang2023federated} aims to learn a model that will be generalized to new domains, but not for personalization. 

\noindent \textbf{Personalized layers.} Given a global model, which layers/components in a network should be personalized to tailor to local distributions attracts increasing attention lately \cite{shen2022cd2,liang2020think,li2021fedbn,bui2019federated,arivazhagan2019federated}. Our goal is orthogonal to this direction since we focus on summarizing participating clients for new clients. For simplicity, we consider all the layers adaptable. Incorporating these techniques to select a partial network to improve further will be our future work. 

\noindent \textbf{Meta-learning.} 
The most relevant approach to ours is meta-learning, which learns a meta-model for rapid personalization for (new) clients~\cite{Khodak2019AdaptiveGM,Chen2018FederatedMW,fallah2020personalized,Jiang2019ImprovingFL}. However, it requires splitting/reusing the training data for meta-validation. Besides, it typically fine-tunes the entire meta-model for each client. Both are not favorable in low data regimes.
Other methods model the relationships between clients \cite{fedfomo,Huang2021PersonalizedCF} for initialization or regularization. The closest work to ours is~\cite{shamsian2021personalized} that summarizes local models into a HyperNetwork~\cite{ha2016hypernetworks}. We provide a more detailed comparison in~\cref{ss_hypernet}.

\emph{Lastly, \FedBasis is inspired by 1) architecture designs in centralized learning that improve a single neural network ~\cite{yang2019condconv,Chen_2020_CVPR,zhang2021basisnet} and 2) the concept of formulating task models as model linear combinations~\cite{evgeniou2007multi}. 
Our novelty is in extending such a concept to PFL through more effective and scalable implementation, identifying difficulties in optimization, and resolving them accordingly.}

\section{Preliminary}

We first provide a short background. In federated learning (FL), the training data are separately collected and stored by $M$ clients. Each client $m\in[M]$ keeps a private set $\sD_m = \{(\vx_i, y_i)\}_{i=1}^{|\sD_m|}$, where $\vx$ is the input (\eg, images) and $y\in\{1,\cdots,C\}=[C]$ is the true label. 

Given the loss function $\ell$ (\eg, cross-entropy) and the empirical risk $\sL_m(\vtheta) = \frac{1}{|\sD_m|}\sum_{(\vx_i, y_i)\in\sD_m} \ell(y_i, h_{\vtheta}(\vx_i))$ of client $m$, where $h_{\vtheta}$ denotes a 
model parameterized by $\vtheta$, \textbf{personalized federated learning (PFL)} aims to learn for each client $m$ a personalized model $\vtheta_m$ tailored to client $m$'s data distribution. While there is no agreed objective function, many existing works~\cite{smith2017federated,li2021ditto,dinh2020personalized,hanzely2020lower,hanzely2020federated,li2019fedmd} solve an optimization problem similar to
\begin{align}
\min_{\{\mOmega, \vtheta_1,\cdots,\vtheta_M\}} \frac{1}{M}\sum_{m=1}^M \sL_m(\vtheta_m) + \sR(\mOmega,\vtheta_1,\cdots,\vtheta_M),
\label{eq:P-obj} 
\end{align}
where $\sR$ is a regularizer to overcome overfitting and $\mOmega$ is its learnable parameter.

Since the training data are decentralized, \cref{eq:P-obj} is typically solved iteratively between local training at the clients and global aggregation at the server (for $\mOmega$) for multiple rounds, inspired by \FedAvg \cite{mcmahan2017communication}.

\paragraph{Challenges in encompassing new clients.} Although solving~\cref{eq:P-obj} can obtain personalized models, it relies on every client to participate in the training. \emph{In reality, not all clients can join the federated training process due to communication or time constraints,} and it remains unclear how to deal with new client $m'\notin [M]$ who arrives \emph{after the federated training is finished}. 
While fine-tuning a pre-trained global model with new client's data can produce a personalized model, it is prone to over-fitting~\cite{li2021ditto,pillutla2022federated,wu2022motley,fallah2020personalized} even with regularization (as will be verified in~\cref{ss:main_exp}). Also, it does not fully leverage the relationships of participating clients, as discussed in \cref{s-intro}.

\section{\FedBasis: PFL with Shareable Bases}
\label{s_approach}

To resolve these issues, we propose a novel personalization approach inspired by \cite{changpinyo2016synthesized,evgeniou2007multi}.
We start with the assumption and formulation, followed by theoretical motivation and implementation.

\subsection{Formulation}
\label{ss_motivation}

 \noindent \textbf{Assumption.} For both participating clients and new clients, the clients' local data share similarity (\eg, domains, styles, classes, etc) --- a common assumption made in multi-task learning~\cite{evgeniou2007multi}. It is likely that we can use a much smaller set of models $\{\vv_1, \dots, \vv_{K}\}$, $K\ll M$, $|\vv| = |\vtheta|$, to construct high-quality personalized models.

\noindent \textbf{Shareable bases.} We represent each personalized model's parameters (\ie, weights) $\vtheta_m$ by a small set of $K$ \emph{basis} models $\sV = \{\vv_1, \dots, \vv_{K}\}$ shared among clients 
\begin{align}
\vtheta_m = \vtheta(\valpha_m, \sV) = \sum_k \valpha_{m}[k] \times \vv_k, \label{e_model_cons}
\end{align}
where $\valpha_{m}\in\Delta^{K-1}$ is a $K$-dimensional vector on the $(K-1)$-simplex, seen as the personalized convex combination coefficients. That is, each personalized model is a convex combination of the basis models. We note that such a combination operation is linear only \emph{within each neural network layer}; the synthesized model is still a neural network with non-linear operations. The representative ability thus remains versatile for constructing personalized models. 

\noindent \textbf{Objective function.} Building upon the model representation in \cref{e_model_cons} and the optimization problem in \cref{eq:P-obj}, we define our \FedBasis PFL problem for learning both the bases $\sV = \{\vv_1, \dots, \vv_{K}\}$ and the coefficients $\sA = \{\valpha_1, \dots, \valpha_M\}$ as\footnote{We drop the regularization term in \cref{eq:P-obj} as the convex combination itself is a form of regularization~\cite{evgeniou2007multi}. We implement $\valpha$ by a softmax function in our experiments.} 
\begin{align}
& \min_{\sA= \{\valpha_m\}_{m=1}^M, \sV = \{\vv_k\}_{k=1}^K} \hspace{2pt} \frac{1}{M}\sum_{m=1}^M \sL_m(\vtheta_m),\nonumber\\
& \quad \text{ where } \vtheta_m = \sum_k \valpha_{m}[k] \times \vv_k.\label{eq:fedbasis}
\end{align}

\noindent \textbf{Training.} We solve~\cref{eq:fedbasis} in a federated setting in~\cref{ss_algorithm}.

\noindent\textbf{Personalization for new clients.} To generate the personalized model, a new client $m'$ receives the learned $\sV$ and finds its specific combination coefficients $\valpha_{m'}$ by SGD with its local data while keeping $\sV$ frozen, \ie, $\valpha_{m'} = \argmin_{\valpha} \sL_{m'}(\vtheta(\valpha, \sV))$ based on~\cref{e_model_cons}. Since $|\valpha_{m'}|$ is mere $K$ per client, it can be robustly learned with fewer data.

\noindent \textbf{Remark.}
We introduce additional advantages of \FedBasis.
First, in inference, \FedBasis enjoys the same memory footprint and computation cost as a single basis model. Convexly combining the parameters (not the predictions!) of the basis models in $\sV$ layer-by-layer according to $\valpha_{m'}$ will merge them into a single personalized model $\vtheta_{m'}$. The inference cost thus remains constant, not scaling with $K$. 

Second, compared to PFL methods based on meta-learning \cite{fallah2020personalized,finn2017model}, which fine-tune the entire or partial model for each (new) client from the meta-learned initialization, we combine the bases \emph{shared} by all clients into a personalized model by learning only a small coefficient vector. This makes \FedBasis more robust to overfitting and hyperparameters when the new client's local data size is small. We will discuss the theoretical benefits of this personalization formulation next in~\cref{ss_hypernet}.

\subsection{Theoretical Motivation}
\label{ss_hypernet}
The theoretical benefit of \emph{summarizing} the clients with fewer \emph{trainable} parameters is outlined by Theorem 1 in~\cite{shamsian2021personalized}, which investigates learning a low-rank approximation over all the personalized model's parameters to reconstruct them. Namely, each personalized model can be represented similarly as in~\cref{e_model_cons}; the theoretical analysis can therefore be applied to \FedBasis. In the following, we briefly review the analysis in the context of \cref{ss_motivation}. The assumptions follow Sec.~4.5 in~\cite{shamsian2021personalized}.

Let $\sV\in\R^{|\vtheta|\times K}$ be the dictionary matrix of total size $Q$, $\sA = [\valpha_1, \dots, \valpha_M]\in\R^{K\times M}$ be the coefficient matrix, and $L$ be the sum of their Lipschitz constants. That is, each client $m$ learns an embedding vector $\valpha_m\in\R^K$ and there are $M$ clients in total. There exists a sample size
\begin{align}
N=\mathcal{O}(\frac{K}{\epsilon^2}\log \frac{L}{\delta} + \frac{Q}{M\epsilon^2}\log \frac{L}{\delta})
\label{eq_bound}
\end{align}
such that if the training samples per client $|\sD_m|>N$, the generalization gap between the true loss and the empirical risk of the personalized model $\vtheta_m:=\vtheta(\valpha_m, \sV)$ will be bounded (\ie, $|\tilde \sL_m(\vtheta_m)-\sL_m(\vtheta_m)|\leq\epsilon$) with probability at least $1-\delta$, for all clients. 
The second term in~\cref{eq_bound} implies that summarizing many clients (a large $M$) with a small dictionary ($Q=|\vtheta|\times K$ with a small $K\ll M$) can notably improve generalization. 

\noindent \textbf{Remark.} In~\cite{shamsian2021personalized}, building on the analysis, the authors proposed to implement $\vtheta_m=\vtheta(\valpha_m, \sV)$ via a multi-layer perception (MLP), with $\valpha_m$ as the input and $\sV$ as the MLP's parameters. In other words, they learned an MLP to predict a neural network $\vtheta_m$, aka a hypernetwork~\cite{ha2016hypernetworks}.
This notoriously increases the size of $\sV$ (\ie, $Q$), making it hard to scale to deeper modern networks. Indeed, in~\cite{shamsian2021personalized}, $Q$ is about $100\times |\vtheta|$ for handling $10\sim100$ clients, which is larger than the sum of clients' model sizes. In contrast, our model representation in~\cref{e_model_cons} exactly follows the formulation in the analysis. 
\FedBasis thus enjoys much fewer parameters --- $Q=K\times|\vtheta|$ with a small $K$ ($4$ to $8$ in our experiments).
Last but not least, \cite{shamsian2021personalized} first learns $\vtheta_m$ and then learns $\valpha_m$ and $\sV$ to reconstruct it. The reconstructed model thus does not necessarily minimize the empirical risk. In contrast, we directly learn the bases and combination coefficients to construct personalized models that minimize the empirical risks, potentially leading to better personalization.

\subsection{Federated Learning Algorithm}
\label{ss_algorithm}

Since \cref{eq:fedbasis} cannot be solved directly, we present an FL algorithm to learn the basis models $\sV$ and the coefficients $\sA$. 
To begin with, we introduce a baseline algorithm via the \FedAvg pipeline, iterating between local and global updates:
\begin{align}
& \text{Local: } &&\hspace{-2pt} \{\valpha_m^{(t)}, \tilde{\sV}_m^{(t)}\} = \argmin_{\{\valpha, \sV\}} \sL_m(\valpha, \sV), \label{eq_baseline_local}\\
& &&\text{ initialized by } \valpha = {\frac{\textbf{1}}{K}} \text{ and } \sV = \bar{\sV}^{(t-1)}, \nonumber \\
& \text{Global: } &&\hspace{-2pt} \bar{\sV}^{(t)} \leftarrow \frac{1}{M}\sum_{m=1}^M{\tilde{\sV}_m^{(t)}},
\label{eq_baseline_glob}
\end{align}
where $\textbf{1}$ is an all-one vector; we use $\sL_m(\valpha, \sV)$ as a concise notation for $\sL_m(\vtheta = \sum_k \valpha[k]\times \vv_k)$. 
In each round $t$, a client first receives the latest $K$ bases $\bar{\sV}^{(t-1)}$ from the server and updates the bases and the coefficient vector\footnote{We note that client $m$ only sees and updates its own combinations $\valpha_m^{(t)}$, not others'. $\valpha_m^{(t)}$ is initialized every round locally and we do not keep it stateful or share it with the server.} by minimizing the local loss $\sL_m$. This results in a local copy of bases $\tilde{\sV}_m^{(t)}$ for client $m$. The global aggregation then returns these $M$ copies of bases to one copy $\bar{\sV}^{(t)}$ by weight averaging \cite{mcmahan2017communication}: the average is taken over the $M$ local copies of each basis.

\noindent\textbf{Problem of bases collapse.} Unfortunately, such naive training can hardly outperform using a single basis, \ie, reducing to a single global model for all clients. To understand this, we investigate the federated training dynamics using a preliminary experiment on the PACS image classification dataset~\cite{PACSli2017deeper} with ResNet18~\cite{he2016deep}, $K=4$ bases, $M=40$, and local epochs $=5$. More details are in the supplementary. We check (1) the average pairwise cosine similarity between the basis model parameters; (2) the average entropy of the learned combination vectors. High entropy implies a more uniform combination vector.
 
\begin{figure}[t]
    \centering
    \minipage{0.45\columnwidth}
    \centering
    \mbox{\small Within one-round local training}
    \includegraphics[width=1\linewidth]{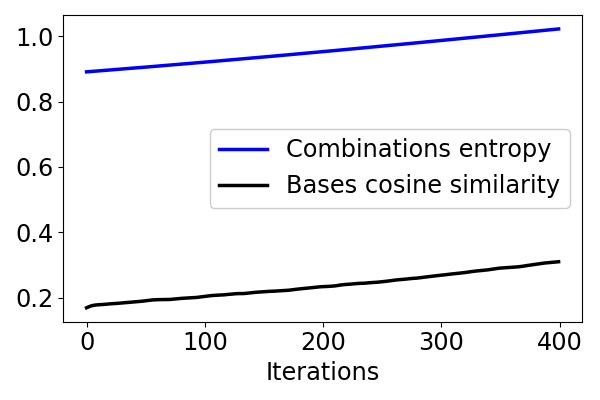} 
    \endminipage
    \hfill
    \minipage{0.45\columnwidth}
    \centering
    \mbox{\small Along training rounds}
    \includegraphics[width=1\linewidth]{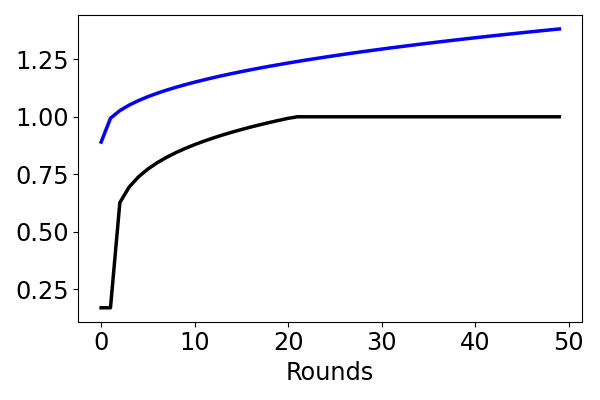}
    \endminipage
    \hspace{0.05cm}
    \vskip -10pt
    \caption{\small \textbf{Cosine similarity between bases, and the entropy of clients' combination vectors on PACS dataset.} \FedBasis by baseline training collapses to non-specialized bases and uniform combinations: (left) within one round of local training for one client; (right) along training rounds, where bases are aggregated at the server and the combinations entropy is averaged over clients.}
    \label{fig:Q3}
    \vskip -10pt
\end{figure}

In~\cref{fig:Q3}, we found that both the pairwise similarity and the entropy increase along with local training SGD iterations and training rounds. In other words, the bases gradually \emph{collapse} to similar parameter values; the combination vectors of all clients nearly collapse to uniform combinations. Consequently, each basis model does not learn specialized knowledge; the whole bases $\sV$ basically degenerate to a single model (or $K$ very similar models). By taking a deeper look at~\cref{fig:Q3}, we found that the collapse problem happens primarily within each local training round. To explain it, let us analyze the gradients derived at local training in \cref{eq_baseline_local},
\begin{align}
    & \nabla_{\vv_k} \sL_m(\valpha, \sV) = && \hspace{-40pt} \valpha[k] \times \nabla_{\vtheta} \sL_m(\vtheta), \nonumber \\
    & \nabla_{\valpha[k]} \sL_m(\valpha, \sV) = && \hspace{-40pt}  \vv_k \cdot \nabla_{\vtheta} \sL_m(\vtheta). 
\end{align}
Interestingly, while with different magnitudes $\valpha[k]$, we see that $\nabla_{\vv_k} \sL_m(\valpha, \sV)$ pushes every local basis model $\vv_k\in\tilde{\sV}_m^{(t)}$ away from the same direction (since $\valpha[k]\geq0$). As local basis models become similar towards $-\nabla_{\vtheta} \sL_m(\vtheta)$, their inner products with $-\nabla_{\vtheta} \sL_m(\vtheta)$ will get larger (\ie, positive) and similar, which would, in turn, push $\valpha[k]$ to be larger via a similar strength. In other words, the more SGD updates we perform within each round of local training, the more similar the local basis models will be and the more uniform the combination coefficients will be. 
\emph{We propose the following treatments to prevent the collapse problem.}

\noindent\textbf{Coordinate descent for the combination coefficients and bases.} Based on the analysis, to prevent the collapse problem, $\valpha$ and $\sV$ should not be updated at the same time. We propose to first update $\valpha$ while freezing $\sV$ and then update $\sV$ while freezing $\valpha$, each for multiple SGD steps, within every local training round. We note that at the beginning of each round of local training, $\vv_k \cdot \nabla_{\vtheta} \sL_m(\vtheta)$ is not necessarily negative. Updating $\valpha$ with frozen $\sV$ thus could potentially enlarge the difference among elements in $\valpha$: forcing the personalized model to attend to a subset of bases. 

\noindent\textbf{Sharpening combination coefficients.} Since $\valpha[k]\geq0$, updating $\vv_k$ locally with $\nabla_{\vv_k} \sL_m(\valpha, \sV)$ would inevitably increase the cosine similarity between basis models. The exception is when some bases get $0$ gradients, \ie, $\valpha[k]=0$. We therefore propose to \emph{artificially} and \emph{temporally} enforce this while calculating $\nabla_{\vv_k} \sL_m(\valpha, \sV)$. We implement $\valpha$ by learning $\vpsi\in\R^K$ and reparameterizing it via a softmax function sharpened with a temperature $0<\tau\leq 1$ as $\valpha[k]=\frac{\exp{(\vpsi[k]/\tau})}{\sum_{k'} \exp{(\vpsi[k']/\tau})}$.

\noindent\textbf{Improved training algorithm.} Putting these treatments together, we present an improved training algorithm for \FedBasis based on~\cref{eq_baseline_local}. Please see the algorithms in the supplementary for the pseudo-code for multi-round training.
\begin{align}
& \text{Local: }\hspace{-4pt} && \text{ initialized by } \valpha = {\frac{\textbf{1}}{K}} \text{ and } \sV = \bar{\sV}^{(t-1)}, & \text{\small[Step 1]} \nonumber \\
& &&\hspace{-2pt} \valpha_m^{(t)} = \argmin_{\valpha} \sL_m(\valpha, \sV), & \text{\small[Step 2]} \nonumber
\\
& &&\hspace{-2pt} \valpha_m^{(t)\dagger} \leftarrow \method{Sharpen}(\valpha_m^{(t)} ; \tau), & \text{\small [Step 3]} \nonumber
\\
&  &&\hspace{-2pt}  \tilde{\sV}_m^{(t)} = \argmin_{\sV} \sL_m(\valpha_m^{(t)\dagger}, \sV), & \text{\small[Step 4]} \nonumber \\
& \text{Global: } \hspace{-4pt} &&\hspace{-2pt} \bar{\sV}^{(t)} \leftarrow \frac{1}{M}\sum_{m=1}^M{\tilde{\sV}_m^{(t)}}. &
\label{eq_improved}
\end{align}

\noindent\textbf{Technical details.} We provide implementation details, including how to initialize each basis in the supplementary. 
 
\noindent\textbf{Computation and communication cost.} While \FedBasis requires more cost in training $K$ models, $K$ is reasonably small and affordable for the modern Internet/GPUs.

\begin{figure}[t]
    \centering
    \includegraphics[width=\columnwidth]{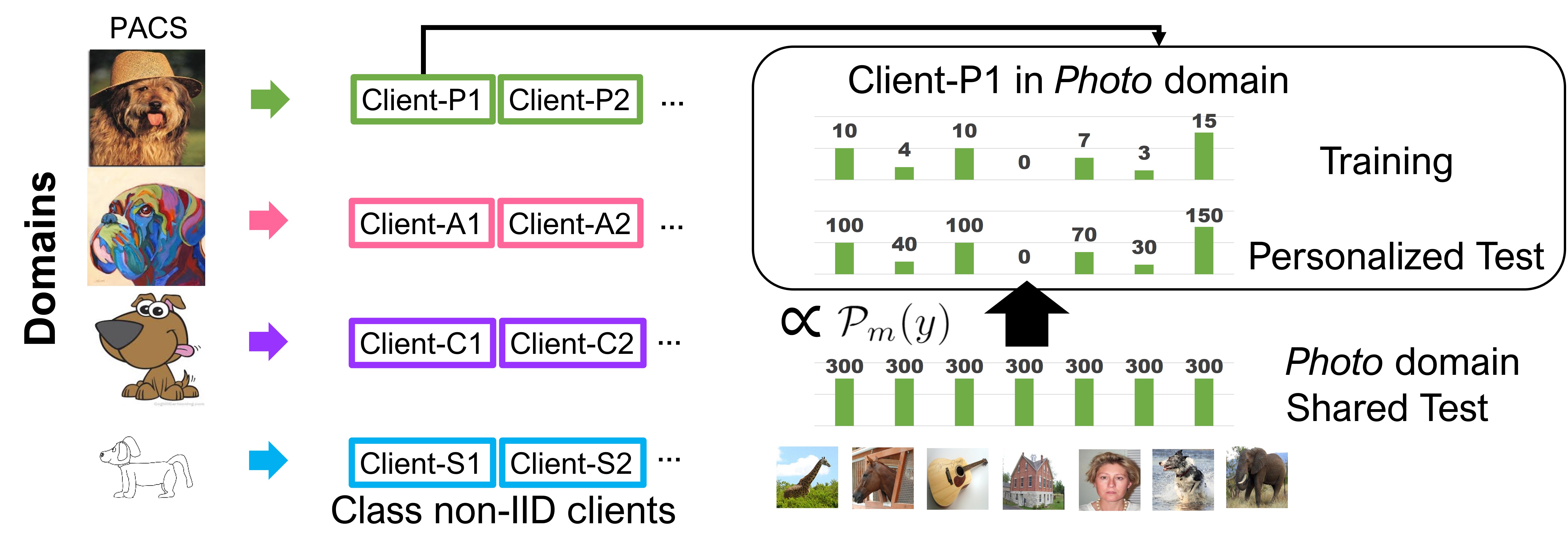}
    \vskip-10pt
    \caption{\small Example of our \PFLBed construction for PFL.}
    \label{fig:dataset}
    \vskip-12pt
\end{figure}

\section{\PFLBed: \emph{bases} for PFL Benchmarks}
\label{s_pflbed}
Many existing efforts are on building \emph{generic} FL datasets~\cite{hsu2020federated,reddi2021adaptive}, including the LEAF benchmarks~\cite{caldas2018leaf} but not for the PFL literature. For the sake of algorithm development, how should we construct a reliable evaluation? As side contributions, we propose the following aspects:

\noindent\paragraph{\vmark Reliable evaluation.} We identify \emph{two challenges} for realistic PFL evaluation on clients each with a small data size. First, \emph{the test sets should be sufficiently large} for statistical reliability. Many previous works~\cite{li2021ditto,shamsian2021personalized,dinh2020personalized} split an even smaller test set for each client. Second, the small local data size can lead to an even more problematic evaluation for realistic non-IID PFL since the \emph{training/test distributions might be mismatched}. For example, the FEMNIST dataset in LEAF benchmark~\cite{caldas2018leaf} on average only has $226$ images over $62$ classes for each writer; many classes only have $\leq1$ images. It is unfaithful to split each client into train/test sets due to mismatches on label distributions $\sP_m(y)$. Indeed, we found a large discrepancy $\frac{1}{M}\sum_m\|\sP^{train}_m(y)-\sP^{test}_m(y)\|_1=0.77$ even with a $50/50 \%$ split. See an illustration in the supplementary. 

\noindent\paragraph{\vmark Cross-domain with non-IID $\sP_m(\vx, y)$.}
A realistic personalized dataset should have the joint distribution $\sP_m(\vx, y)$ differ from client to client, not just $\sP_m(\vx)$ (\eg, domains~\cite{li2021fedbn}) or $\sP_m(y)$ (\ie, class labels~\cite{collins2021exploiting,fallah2020personalized,shamsian2021personalized}). Both the training data sizes and the class distributions should be skewed among clients to simulate realistic cases.  

To achieve these desired properties for PFL training and evaluation, we propose to transform a cross-domain dataset $\sD$ that each input is associated with a domain annotation, into clients' sets $\{(\sD^{train}_m, \sD^{test/val}_m)\}$ with the following procedures, as illustrated in~\cref{fig:dataset}:
\begin{enumerate}[noitemsep,topsep=0pt,parsep=0pt,partopsep=0pt,leftmargin=*]
\item Separate $\sD$ based on its domain annotations.
\item For each domain, split the class-balanced test and validation sets which will later be shared with all clients from this domain. Take the rest as the training set.
\item For the training set per domain, create a class-heterogeneous partition, \eg, by the commonly used Dirichlet sampling~\cite{hsu2019measuring} for $M'$ clients. Each client's images are class-non-IID and from a single domain.

\item For each client $m$, record the class distributions $\sP_m(y)$ of its training set.
\item For each client in each domain, assign the whole test set of the same domain as $\sD^{test/val}_m$. 

\item Compute $\frac{1}{M}\sum_m\frac{\sum_i\sP_m(y_i)\textbf{1}(y_i = \hat y_{i})}{\sum_i\sP_m(y_i)}$ as the client-wise average personalized accuracy during evaluation.

\end{enumerate}

\noindent\paragraph{Evaluation on new clients.} To evaluate how practical a personalized system can serve new clients, one can split the clients into participating/new clients groups and train on the participating group. After training, personalization is performed on each new client's training set $\sD^{train}_m$, and follows the same testing protocol.

\begin{table}
\caption{\small Summary of the datasets and setups.}
\vskip-7pt
\centering
\small 
\setlength{\tabcolsep}{2pt}
\renewcommand{\arraystretch}{0.45}
\begin{tabular}{l|ccccc}
\toprule
Dataset & Size & $\#$Class & Resolution & Domain & Client Split\\
\midrule
PACS & 9K & 7 &224$^2$ & Styles & \PFLBed \\
Office-Home & 16K & 65 & 224$^2$ & Styles & \PFLBed \\
GLD23K & 23K & 203 & 224$^2$ & Natural & User ID \\
CIFAR-10/100 & 60K & 10/100 & 32$^2$ & None & Dirichlet \\
\bottomrule
\end{tabular}
\label{tbl:dataset}
\vskip -15pt
\end{table}

\noindent\textbf{Examples.}
We consider the two image object recognition datasets PACS~\cite{PACSli2017deeper} and Office-Home~\cite{OHvenkateswara2017deep} that are widely used in domain adaptation, both providing $4$ domain annotations of image \emph{styles}. Following the proposed \PFLBed procedures, we first split the samples of each domain into $60$/$20$/$5$/$15 \%$ for training, new, validation, and test sets. The training/new sets are further split for $20/10$ of participating/new clients per domain by class non-IID sampling with Dirichlet($0.3$), following~\cite{hsu2019measuring}.

\newcolumntype{a}{>{\columncolor{green!20}}c} 

\begin{table*}[t]
    \caption{\small Personalized test accuracy ($\%$) on non-IID new clients. Averaged over 3 runs. Variances and more results are provided in the supplementary. Each method personalizes on each client's local data of different sizes. We personalize by fine-tuning (FT) for $20$ epochs and record at the \textbf{Last} and \textbf{Best} (by validation) epoch, and measure the difference {\textbf{$|\Delta|$}} as personalization robustness.}
    \vskip -5pt
    \label{tab:noniid}
    \centering
    \small
    \setlength{\tabcolsep}{1.5pt}
	\renewcommand{\arraystretch}{0.7}
    \begin{tabular}{ll|cc|a|cc|a|cc|a|cc|a|cc|a|cc|a|cc|a}
    \toprule 
    \multicolumn{2}{c|}{Dataset (Part./New Clients)} & \multicolumn{6}{c}{PACS (80/40)} & \multicolumn{6}{c}{Office-Home (80/40)} & \multicolumn{9}{c}{GLD23k (117/116)}\\
    \midrule
    \multicolumn{2}{c|}{Local Size for Personalization}  & \multicolumn{3}{c}{\textbf{S}} & \multicolumn{3}{c|}{\textbf{M}} & \multicolumn{3}{c}{\textbf{S}} & \multicolumn{3}{c|}{\textbf{M}} & \multicolumn{3}{c}{\textbf{S}} & \multicolumn{3}{c}{\textbf{M}} & \multicolumn{3}{c}{\textbf{L}}\\
    \midrule
    \multicolumn{2}{c|}{Approach / Stopping Epoch} & Last & Best & $|\Delta|$ & Last & Best &$|\Delta|$ & Last & Best &$|\Delta|$ & Last & Best &$|\Delta|$ & Last & Best &$|\Delta|$ & Last & Best &$|\Delta|$ & Last & Best &$|\Delta|$\\
    \midrule
    \midrule
    \multirow{2}{*}{\parbox{1.5cm}{\centering \scriptsize Personalized Layers}} & \FedRep  & 87.4 & 87.4 & 0.0 & 92.5 & 92.4 & 0.1 & 75.6 & 75.6 & 0.0 & 76.0 & 76.1 & 0.1 &75.7 & 77.6 & 1.9 & 78.8 & 78.8 & 0.0 & 80.1 & 80.8 & 0.7\\
    & \FedBN  & 86.2 & 88.2 & 2.0 & 92.4 & 92.4 & 0.0 & 76.9 & 77.0 & 0.1 & 78.1 & 78.1 & 0.0 & 74.1 & 74.5 & 0.4 & 76.6 & 76.6 & 0.0 & 76.4 & 76.5 & 0.1\\
    \midrule
     \multirow{3}{*}{\parbox{1.5cm}{\centering \scriptsize Meta-Model}} & \pFedHN &  \multicolumn{2}{c|}{85.4} & - & \multicolumn{2}{c|}{85.5} & - & \multicolumn{2}{c|}{74.1} & - & \multicolumn{2}{c|}{74.3} & - & \multicolumn{2}{c|}{74.5} & - & \multicolumn{2}{c|}{75.6} & - & \multicolumn{2}{c|}{77.2} & -\\
    & \pFedHNFT & 90.5 & 91.2 & 0.7 & 90.4 & 91.4 & 1.0 & 76.2 & 77.2 & 1.0 & 77.1 & 77.6 & 0.5 & 77.6 & 81.4 & 3.8 & 78.6 & 81.6 & 3.0 & 80.2 & 82.2 & 2.0\\    
    & \PerFedAvgFT  & \textbf{95.4} & \textbf{95.6} & 0.2 & \textbf{96.2} & \textbf{96.3} & 0.1 & 84.3 & 84.4 & 0.1 & 86.1 & 86.2 & 0.1 & 78.5 & 85.3 & 6.8 & 79.9 & 85.2 & 5.3 & 82.2 & 86.1 & 3.9\\
    \midrule   
    \multirow{4}{*}{\parbox{1.5cm}{\centering \scriptsize General Representation}} & \kNNPer  & \multicolumn{2}{c|}{71.6} & - & \multicolumn{2}{c|}{71.6} & - &  \multicolumn{2}{c|}{50.4} & - & \multicolumn{2}{c|}{54.5} & - & \multicolumn{2}{c|}{54.0} & - & \multicolumn{2}{c|}{57.4} & - & \multicolumn{2}{c|}{69.2} & -\\
    & \kNNPerFT & 72.7 & 72.7 & 0.0 & 79.4 & 79.7 & 0.3 & 51.6 & 52.4 & 0.8 & 54.2 & 54.4 & 0.2 & 54.2 &54.5 & 0.3 & 57.1 & 57.8 & 0.7 & 69.5 & 70.2 & 0.7\\  
    & \FedAvg & \multicolumn{2}{c|}{88.1} & - & \multicolumn{2}{c|}{88.1} & - & \multicolumn{2}{c|}{73.1} & - & \multicolumn{2}{c|}{73.1} & - & \multicolumn{2}{c|}{45.4} & - & \multicolumn{2}{c|}{45.4} & - & \multicolumn{2}{c|}{45.4} & -\\
    & \FedAvgFT  & 86.1 & 91.9 & 5.8 & 90.5 & 90.5 & 0.0 & 76.1 & 77.4 & 1.3 & 78.2 & 78.5 & 0.3 & 81.5 & 84.2 & 2.7 & 81.6 & 84.5 & 2.9 & 84.1 & 86.1 & 2.0\\
    \midrule     
    \multirow{1}{*}{\parbox{1.5cm}{\centering \scriptsize Ours}} & \FedBasis & 95.2 & 95.2 & 0.0 & \textbf{96.2} & 96.2 & 0.0 & \textbf{87.4} & \textbf{87.5} & 0.1 & \textbf{87.5} & \textbf{87.7} & 0.2 & \textbf{87.4} & \textbf{87.4} & 0.0 & \textbf{87.6} & \textbf{87.6} & 0.0 & \textbf{89.0} & \textbf{89.1} & 0.1\\
    \bottomrule
    \end{tabular}
    \vskip -3pt
\end{table*}

\begin{figure*}[t]
    \centering
    \minipage{0.35\linewidth}
    \centering
    \includegraphics[width=1\linewidth, height=3.5cm]{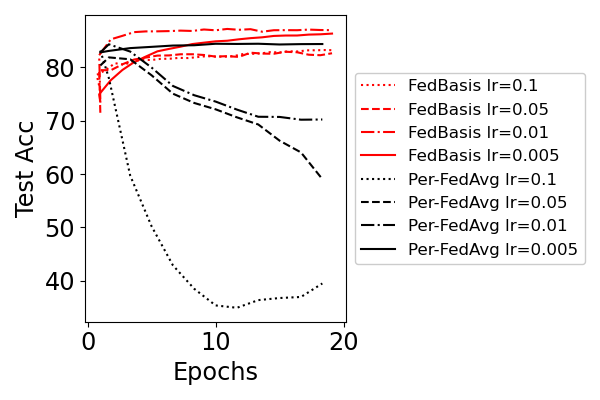} 
    \vskip -5pt
    \mbox{\small (a) Fine-tuning on Office-Home (Small)}
    \endminipage
    \hfill
    \minipage{0.65\linewidth}
    \centering
    \includegraphics[width=1\linewidth]{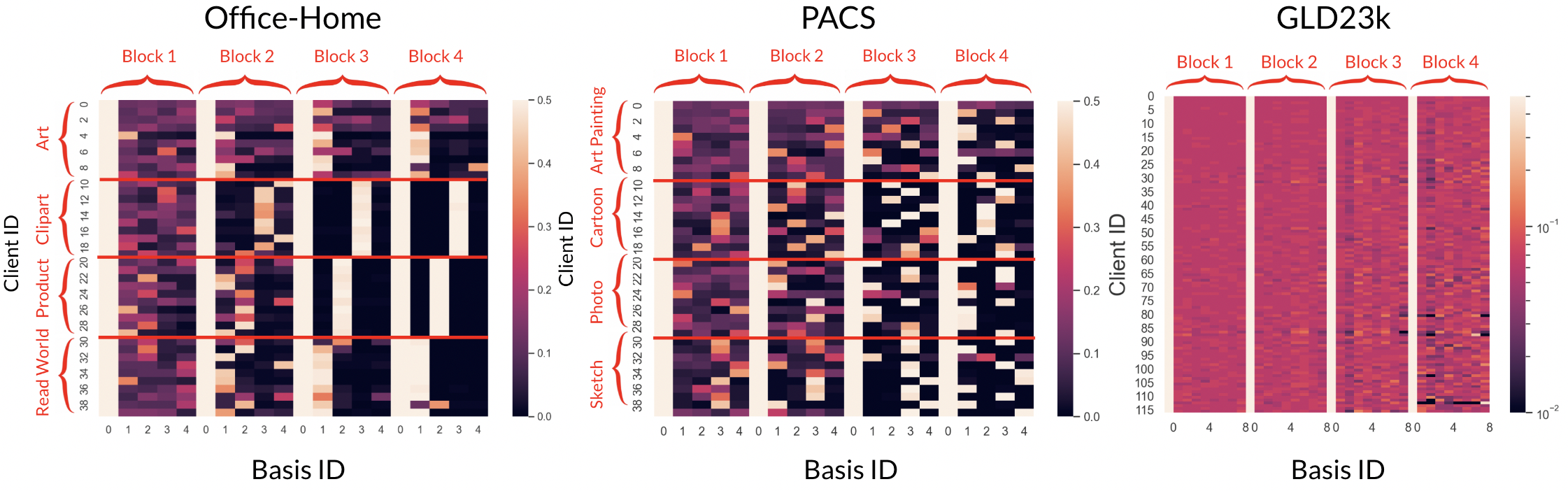}
    \vskip -5pt
    \mbox{\small (b) Learned basis combinations (a combination per ResNet block)}
    \endminipage
    \vskip -5pt
    \caption{\small \textbf{(a) \FedBasis for more robust personalization.} Fine-tuning curves with different learning rates. \textbf{(b) Visualization of $\{\valpha_m\}$.}} 
    \vskip -5pt
    \label{fig:vis}
\end{figure*} 
\section{Experiments}
\label{s_exp}

\subsection{Settings (see the supplementary for details)}

\noindent\paragraph{Dataset.} Besides \PFLBed, for completeness, we include a naturally non-IID Google Landmark (GLD-v2)~\cite{weyand2020google} dataset that has $233$ photographers as clients~\cite{hsu2020federated}. We also consider standard PFL setups using CIFAR datasets~\cite{krizhevsky2009learning} (see~\cref{ss:disc}). \cref{tbl:dataset} summarizes the statistics. 

\noindent\paragraph{Baselines.}
We compare \FedBasis to the state-of-the-art approaches discussed in~\cref{s_related}. \emph{(1) Personalized layers} fine-tunes only classifiers (\FedRep~\cite{collins2021exploiting}) or batchnorm parameters (\FedBN~\cite{li2021fedbn}) for new clients. The most relevant approach to ours summarizes clients into a \emph{(2) meta-model}, including \pFedHN~\cite{shamsian2021personalized} based on hypernetworks that generate a model for each client by learning an input embedding for the hypernetwork.  Another method \PerFedAvg~\cite{fallah2020personalized}\footnote{We focus on the better first-order version and we have compared it with the Hessian-free version in the supplementary.} is based on MAML that learns a good initial model for fine-tuning. As pointed out by~\cite{yu2020salvaging,Wang2019FederatedEO,chen2022bridging,cheng2021fine}, fine-tuning (\textbf{FT}) on \emph{(3) general representations} from global model like \FedAvg~\cite{mcmahan2017communication} serves as a strong baseline. \kNNPer~\cite{marfoq2022personalized} further builds $k$-nearest neighbors classifier on top of the features locally. 

\noindent\paragraph{\FedBasis.} We train with $5$ local epochs for both $\valpha$ and $\sV$ as described in~\cref{ss_algorithm}, where $\tau=0.1$ for sharpening the combinations and each ResNet block uses its own combination vector. The number of bases is $4/4/8$ for PACS/Office-Home/GLD. For personalization, only the combinations and classifier are trained with $\sV$ frozen. 

\noindent\paragraph{FL Training.} ImageNet pre-trained ResNet-18~\cite{he2016deep} with standard pre-processing is used with SGD optimizer with $0.9$ momentum, $1\text{e-}4$ weight decay, and $0.01$ local learning rate. \PFLBed/GLD datasets are trained for $100/200$ rounds with $16/64$ batch sizes and $5$ local epochs (sample $100/10\%$ participated clients) in each round. 

\noindent\paragraph{Personalization.} ``New clients'' are adapted with different local data sizes (\textbf{S}mall/\textbf{M}oderate/\textbf{L}arge) with the learning rate tuned from $\{0.005, 0.01, 0.05\}$ and $1\text{e-}4$ weight decay.

\subsection{Main Results} 
\label{ss:main_exp}

We highlight the following observations in~\cref{tab:noniid}: 

\noindent\textbf{\vmark Meta models are promising.}
The best baseline is \PerFedAvgFT, supporting that modeling personalization from a meta-view is promising since it considers the inter-client relationships. Our \FedBasis summarizes bases over clients but with fewer trainable parameters per client, thus leading to more robust personalization and outperforming the baselines, especially on harder datasets Office-Home and GLD, supporting our~\cref{ss_hypernet}. 

\noindent\textbf{\vmark  Fine-tuning the feature extractor helps.} Fine-tuning on general features (\eg, \FedAvg's global model) helps the performance, validating that the features are preferred to be client-specific. \FedAvgFT is competitive against more recent methods like \FedRep and \kNNPer\footnote{Interestingly, \kNNPer seems to be less effective in such low-data regimes, consistent with~\cite{marfoq2022personalized}. We were able to reproduce the original results where each client has more samples.}.

\noindent\textbf{\vmark  Fine-tuning can be vulnerable w/o careful validation.} However, fine-tuning can be unstable w.r.t. the tuned epochs or likely suffer from overfitting (up to $7\%$ of {\textbf{$|\Delta|$}}), especially when the local size is small. Note that, selecting the \emph{best} epoch is not always feasible  since the clients may not have enough data for validation~\cite{wu2022motley}; thus such robustness is important in practice. 

\noindent\textbf{\vmark \FedBasis is both robust and accurate.} \FedBasis can personalize the whole model by the layer composition ability enforced in training while being robust since it learns much fewer parameters per client. To see it from another view, we further compare the best baseline \PerFedAvgFT with different learning rates. In~\cref{fig:vis} (a), we observe \FedBasis is clearly more robust to hyperparameters such as the learning rates and stopping epoch.

\noindent\textbf{\vmark Regularized personalization is not enough.}
One might wonder if adding a regularizer to fine-tuning can help. We compare a regularizer-based method \pFedMe~\cite{dinh2020personalized} and apply its Eq. (2) for \PerFedAvg. As~\cref{tab:reg} shows, better regularizers can improve slightly but \FedBasis still outperforms.

\begin{table}[t]
    \caption{\small Office-Home new clients regularized fine-tuning (FT).}
    \vskip -5pt
    \label{tab:reg}
    \centering
    \small
    \setlength{\tabcolsep}{1.5pt}
	\renewcommand{\arraystretch}{1}
    \begin{tabular}{cc|c}
    \toprule
    FL Training & FT Regularizer & Acc.\\
    \midrule
    \multirow{1}{*}{\PerFedAvg} & Weight decay / Eq. (2) in \pFedMe & 84.3 / 84.5\\
    \multirow{1}{*}{\pFedMe} & Weight decay / Eq. (2) in \pFedMe & 77.5 / 77.8\\
    \midrule
    \FedBasis & Weight decay & 87.5\\
    \bottomrule
    \end{tabular}
\end{table}

\begin{table}
    \caption{\small Ablation studies of \FedBasis designs in~\cref{s_approach}.}
    \vskip -5pt
    \label{tab:abl}
    \centering
    \small
    \setlength{\tabcolsep}{1.5pt}
	\renewcommand{\arraystretch}{0.65}
    \begin{tabular}{ccc}
    \toprule
    \texttt{CoordinateDescent}  & $\tau$ & Office-Home Acc.\\
    \midrule
    \xmark &   0.1 & 83.5 \\
    \cmark &   1.0 & 87.1 \\
    \cmark &   0.1 & 87.5 \\
    \bottomrule
    \end{tabular}
    \vskip -10pt
\end{table}

\subsection{Further Discussions} 
\label{ss:disc}
\noindent\textbf{Visualization.}
To understand \FedBasis, we visualize the learned combinations in~\cref{fig:vis} (b). Interestingly, we see the clients group together according to domains especially in the latter blocks (\eg, Office-Home Block 3 \& 4). \FedBasis enables the shareable bases to automatically determine the collaboration among the non-IID clients.  

\noindent\textbf{Ablations.} The ablation study (w/ moderate local size) in~\cref{tab:abl} verifies our designs in~\cref{s_approach}. 

\noindent\textbf{Sanity checks: conventional PFL setups on CIFAR.}
So far the main study is on our \PFLBed setups, we provide two sets of standard PFL experiments on the CIFAR-10/100 benchmarks\footnote{Compared to \PFLBed in~\cref{s_pflbed}, these simulated datasets are single-domain and purely class non-IID. The evaluation is faithful since the training/test sets are distributionally matched.} for proving the generalizability of \FedBasis. We use $3$ bases for CIFAR experiments. 

First, we use the authors’ official codes to reproduce and compare Table 1 experiments in~\cite{collins2021exploiting} of participating clients, including the backbone and training detail.~\cref{tbl:cifar} shows \FedBasis is also effective in this setup and performs comparably to the state of the arts like \FedRep based on global features. We note that this is expected since the images are single-domain thus the features might not have much room to be personalized. This can be seen in the saturated improvements that many PFL algorithms perform similarly to \FedAvgFT in this class non-IID setting. In contrast, \PFLBed in~\cref{s_pflbed} is cross-domain and class non-IID where fine-tuning features is important.

Second, we further evaluate new client personalization as our main goal in~\cref{tab:noniid}. We follow~\cref{tbl:cifar} but now split the 100 clients into 80/20 clients for training and evaluation. Each new client is fine-tuned for 10 epochs. In~\cref{tbl:new_cifar}, we see our \FedBasis is competitive against the strongest baseline (\PerFedAvgFT).

\begin{table}[t]
\caption{\small Test accuracy of \emph{participating} clients on conventional PFL of CIFAR ($100$ clients, each has $5$ classes). We \emph{reproduce} the setting and the baseline results of Table 1 in~\cite{collins2021exploiting}.}
\begin{center}   
\small
\setlength{\tabcolsep}{1pt}
\renewcommand{\arraystretch}{0.8}
\vskip -10pt
\begin{tabular}{l|c|c}
\toprule
Method & CIFAR10 & CIFAR100 \\
\midrule
FedAvg~\cite{mcmahan2017communication}/ +FT & 51.8/73.7  & 23.9/79.3 \\ 
FedProx~\cite{li2020federated}/ +FT  & 51.0/72.8  & 20.2/78.5 \\
\cite{karimireddy2020scaffold}/ +FT  & 47.3/68.2  &  20.3/78.9 \\
\midrule 
Fed-MTL \cite{smith2017federated}  & 58.3   & 71.5 \\
Per-FedAvg (Fallah et al. 2020)  & 67.2 & 72.1 \\
LG-Fed \cite{liang2020think}   & 63.0 & 72.4 \\
L2GD \cite{hanzely2020federated}  & 60.0  & 72.1  \\
APFL  \cite{deng2020adaptive}  & 72.2  & 78.2  \\
Ditto \cite{li2021ditto}  & 70.3  & 78.9  \\
FedPer \cite{arivazhagan2019federated}  & 73.8  & 76.0 \\
FedRep~\cite{collins2021exploiting}   & 75.7  & 79.1 \\ 
\midrule
FedBasis & 75.5 $\pm$ 0.7 & 80.8 $\pm$ 0.5 \\
\bottomrule
\end{tabular}
\end{center}
\vskip -5pt
\label{tbl:cifar}
\end{table}

\begin{table}[t]
\caption{\small CIFAR personalization on \emph{new} clients. $80/20$ participating /new clients, each has $5$ classes.}
\begin{center}
\small
\setlength{\tabcolsep}{3.5pt}
\renewcommand{\arraystretch}{0.75}
\vskip -7pt
\begin{tabular}{l|c|c}
\toprule
Method & CIFAR10 & CIFAR100 \\
\midrule
\FedAvgFT & 72.7 $\pm$ 0.3  & 78.1 $\pm$ 0.3 \\
\PerFedAvgFT & 74.8 $\pm$ 0.4  & 78.4 $\pm$ 0.4 \\ 
\midrule
\FedBasis & 76.5 $\pm$ 0.6 & 79.2 $\pm$ 0.4 \\
\bottomrule
\end{tabular}
\end{center}
\vskip -20pt
\label{tbl:new_cifar}
\end{table}

\section{Conclusion}

We present a novel framework called \FedBasis for robust personalization of new clients. \FedBasis synthesizes personalized models using a few shareable basis models learned from participating clients in federated training. This reduces the learnable parameter size for each client for personalization and mitigates the vulnerability of fine-tuning. We design our federated algorithm to overcome the difficulty in optimization systematically.
We also present a carefully designed benchmark \PFLBed to support future research. We discuss limitations and future work in the supplementary. 

\section*{Acknowledgments}
\label{suppl-sec:ack}
This research is supported in part by grants from the National Science Foundation (IIS-2107077, OAC-2118240, and OAC-2112606) and Cisco Research. We are thankful for the generous support of the computational resources by the Ohio Supercomputer Center.

\bibliography{main}

\clearpage

\appendix
\onecolumn
\begin{center}
\textbf{\Large Supplementary Materials}
\end{center}

\renewcommand{\thesection}{\Alph{section}}
\setcounter{table}{0} 
\renewcommand{\thetable}{\Alph{table}}
\setcounter{figure}{0} 
\renewcommand{\thefigure}{\Alph{figure}}
\setcounter{equation}{0} 
\renewcommand{\theequation}{\Alph{equation}}

 We provide the details omitted in the main paper. 
\begin{itemize}
    \item \cref{suppl-sec:alg}: pseudo codes and more discussion of \FedBasis  (cf. \cref{s_approach} of the main paper).
    \item \cref{suppl-sec:exp}: additional details of experiment setups (cf. \cref{s_approach} and \cref{s_exp} of the main paper). 
    \item \cref{suppl-sec:dataset}: additional discussion on the datasets and \PFLBed (cf. \cref{s_pflbed} of the main paper).
    \item \cref{suppl-sec:results}: additional results and discussion (cf. \cref{s_approach} and \cref{s_exp} of the main paper, including the bases collapse problem in~\cref{ss_algorithm}). 
    
\end{itemize}

\section{\FedBasis Algorithm}
\label{suppl-sec:alg}

\begin{algorithm}[H]
\SetAlgoLined
\caption{\FedBasis --- federated training for the bases}
\label{alg:fedbasis}
\SetKwInOut{Input}{Input}
\SetKwInOut{SInput}{Server input}
\SetKwInOut{CInput}{Client $m$'s input}
\SetKwInOut{SOutput}{Server output}
\SetKwInOut{COutput}{Client $m$'s output}
\SInput{initial global basis parameter $\bar{\sV}$;}
\CInput{local loss $\sL_m$, temperature $\tau$;} 
\For{$t\leftarrow 1$ \KwTo $T$ rounds}{
{\textbf{Communicate} $\bar{\sV}$ to all clients $m\in [M]$;}\\
\For{each client $m\in [M]$ in parallel}{
{\textbf{Initialize} $\{\valpha_m, \sV\} \text{ by } \{\vct{\frac{1}{K}}, \bar{\sV}\}$;}\\
$\valpha_m^\star = \argmin_{\valpha_m} \sL_m(\valpha_m, \sV)$ \\
$\valpha_m^{\star\dagger} \leftarrow \method{Sharpen}(\valpha_m^\star ; \tau)$; \\
$\sV_m^{\star} = \argmin_{\sV} \sL_m(\valpha_m^{\star\dagger}, \sV)$; \\
{\textbf{Communicate} $\sV_m^{\star}$ to the server;}\\
}
\textbf{Construct} $\bar \sV =\frac{1}{M}\sum_{m=1}^M \sV^\star_m$;\\
}
\SOutput{$\bar \sV$;}
\end{algorithm}

\begin{algorithm}[H]
\SetAlgoLined
\caption{\FedBasis --- generate a personalized model}
\label{alg:fedbasis_inf}
\SetKwInOut{Input}{Input}
\SetKwInOut{CInput}{Client $m$'s input}
\SetKwInOut{COutput}{Client $m$'s output}
\CInput{initial global basis parameter $\sV$, local loss $\sL_m$;} 
{\textbf{Initialize} $\valpha_m \text{ by } \vct{\frac{1}{K}}$;}\\
$\valpha_m^\star = \argmin_{\valpha_m} \sL_m(\valpha_m, \sV)$ \\
\textbf{Construct} $\vtheta_m(\valpha_m, \sV) = \sum_k \valpha_{m}[k] \times \vv_k$;\\
\COutput{$\vtheta_m$;}
\end{algorithm}

We provide a summary in~\cref{alg:fedbasis} for training our \FedBasis (cf.~\cref{ss_algorithm} in the main paper) and~\cref{alg:fedbasis_inf} shows how to use it for generating a personalized model. Similar to the \FedAvg algorithm, our \FedBasis also executes a multi-round training procedure between the local training at the clients and aggregation at the server.  

The goal of \FedBasis is to collaboratively train $K$ basis models $\sV=\{\vv_k\}_{k=1}^K$ which can be used to combine into personalized models based on each client's combination coefficient $\valpha_m \in \R^K$ (or more specifically, $\Delta^{(K-1)}$; see~\cref{e_model_cons}) within limited $T$ rounds of communications. The parameters are linearly combined layer by layer. Such specialized layers improve the performance with little extra inference cost. Our contribution is to extend such concepts to personalization in FL setting, identify optimization issues, and resolve them. 

To effectively learn the bases for personalization, in~\cref{ss_algorithm}, we introduce several important techniques in the local training to avoid bases collapse and encourage each basis to learn specialized knowledge. 
In each round of local training at a client $m$, it first initializes the bases $\sV$ using the one broadcast by the server.  Next, we train $\valpha_m$ and $\sV$ with coordinate descent. We update $\valpha_m$ (for multiple SGD steps) while freezing $\sV$ (line $5$ in~\cref{alg:fedbasis}). To force the personalized model to attend to a subset of bases, we sharpen $\valpha_m$ by injecting a temperature into the Softmax function (line $6$ in~\cref{alg:fedbasis}). Then, we update $\sV$ (for multiple SGD steps) while freezing $\valpha_m$. Finally, the updated bases are sent back to the server for a basis-wise average with other clients' updates.    

The \FedBasis formulation enjoys several desired properties.
\begin{itemize} [topsep=1pt] 
    \item The total learnable parameter size of all the personalized models (almost) does not scale with the number of clients. \FedBasis ultimately outputs the bases $\sV$ with combination coefficients $\valpha_m$ for each client $m$. Each client only has $|\valpha_m|=K$ personalized parameters, which is negligible compared to the model. After $\sV$ is trained, it can be used to generate personalized models for new clients. As discussed in~\cref{ss_hypernet}, such formulation leads to more generalized personalization and robustness to small data sizes.
    \item For local training, the combined model only needs to be generated per mini-batch but not per instance, making it scalable to batch sizes. 
    \item The size of communications is $K$ times more but $K$ is typically small ($4\sim8$ in our experiments).   
    \item \FedBasis does not increase clients' computation cost in inference. After training, the basis models are combined into a single personalized model for each client. This is sharply different from approaches based on the mixture of models in that input needs to go through every expert and ensembles the predictions, where the cost is linear to the number of experts.
\end{itemize}

\section{More experimental details}
\label{suppl-sec:exp}
\subsection{Split new clients for evaluation}
Following the proposed \PFLBed procedures, we first split the samples of each domain into $60$/$20$/$5$/$15 \%$ for training, new, validation, and test sets. The training/new sets are further split for $20/10$ of participating/new clients per domain by class non-IID sampling with Dirichlet($0.3$), following~\cite{hsu2019measuring}. 
In our experiments, to demonstrate the data efficiency of each of the methods, we consider different training sizes (\textbf{S}mall/\textbf{M}oderate/\textbf{L}arge) for personalization of each client. Concretely, for Office-Home and PACS, we use $50$/$100\%$ of each client's training set as the \textbf{S}/\textbf{M} setting for personalization, respectively. We note that, in \PFLBed, we already split a relatively small set ($20\%$ of the overall data) and further split it into several new clients. On the other hand, for the GLD-v2 dataset, the clients are already split by User IDs, we thus randomly split $10$/$20$/$40\%$ of each new client's data as the training set and take the rest as the test/validation sets (we split $20\%$ for validation).

\subsection{Technical Extension}
\label{ss-extensino}
\noindent\textbf{Block-wise combinations.}
In~\cref{e_model_cons}, it applies the same coefficient $\valpha_m[k]$ to combine the whole $\vv_k$ into $\vtheta_m$ .
Such a formula can be slightly relaxed to decouple the coefficients by layers, allowing it to learn different collaboration patterns. For instance, in our experiments on ResNets, we learn a coefficient vector for each of the $4$ blocks and the classifier (instead for the whole network). 

\noindent\textbf{The major basis and warm-start for the bases.}
One concern is that an individual basis can be specialized but poorly generalized since it is likely trained on only partial data. We show this can be resolved easily with two tricks. First, we maintain a \emph{major basis} that is always included in the combinations. That is, \cref{e_model_cons} becomes $\vtheta_m(\valpha_m, \sV) = \frac{1}{2}(\vv' + \sum_k \valpha_{m}[k] \times \vv_k$), where $\vv'$ is the major basis and the other bases personalize on top of it. Second, \FedBasis can be a post-processing tool for a generic FL algorithm for personalizing new clients. Practically, we first run \FedAvg for a few rounds. The server collects the local models $\{\vtheta_m\}$, clusters them into $K$ clusters, and initializes $K$ basis models with the centroids. It warm-starts \FedBasis since each basis already learns general knowledge and is somehow specialized. We run \FedAvg for $30\%$ of the total rounds and collect its global/local models~\cite{chen2022bridging} to warm-start the major/non-major bases, respectively.

\subsection{Hyperparameters}
\label{suppl-sec:hyper}
For every method, we first conduct federated training, then personalize the trained model for further personalization on new clients as evaluation. 

\noindent\paragraph{Training.} We use an ImageNet pre-trained ResNet-18~\cite{he2016deep} with standard ImageNet-style pre-processing, SGD optimizer with $0.9$ momentum, $1\text{e-}4$ weight decay, and $0.01$ local learning rate. \PFLBed/GLD datasets are trained for $100/200$ rounds with $16/64$ batch sizes and $5$ local epochs for each participating client (sample $100/10\%$) in each round. All the methods including the baselines and ours use the same training process for a fair comparison. For \FedBasis, it is trained with $5$ local epochs for both $\valpha$ and $\sV$ as described in cf.~\cref{ss_algorithm} and~\cref{ss-extensino}, where $\tau=0.1$ for sharpening the combinations. The number of bases is $4/4/8$ besides the major basis for PACS/Office-Home/GLD. 

\noindent\paragraph{Evaluation.} We consider that ``new clients'' are personalized with different local data sizes (\textbf{S}mall/\textbf{M}oderate/\textbf{L}arge) with the learning rate tuned from $\{0.005, 0.01, 0.05\}$ and $1\text{e-}4$ weight decay. Each method is personalized in the way they proposed. We further consider different strategies including linear probes and fine-tuning and summarize in~\cref{sup-tab:noniid_more} for completeness. 

Following the same personalization adaptation, each method is trained for each client to produce its personalized model. For the personalized layers approach, we first train by their algorithms, then personalize those layers for new clients like classifiers (\FedRep~\cite{collins2021exploiting}) or batchnorm parameters (\FedBN~\cite{li2021fedbn}). \kNNPer~\cite{marfoq2022personalized} uses the global features of \FedAvg~\cite{mcmahan2017communication} for $k$-nearest neighbors based classification locally. For \pFedHN~\cite{shamsian2021personalized}, it first trains a hypernetwork that is a model generator. We follow~\cite{shamsian2021personalized} to train each client an input embedding for the hypernetwork to generate its personalized model (which can be further fine-tuned fully). \PerFedAvg~\cite{fallah2020personalized} is based on MAML that learns a good initial model for fine-tuning. Therefore, it should not be used directly. We consider using full fine-tuning on it. We focus on the better first-order version and we have compared it with the Hessian-free version in~\cref{tab:pfedavvg}. For personalization with \FedBasis, only the combinations and classifier are trained.

\section{More details on the \PFLBed dataset construction}
\label{suppl-sec:dataset}

\subsection{Discussions on \PFLBed}
\label{suppl-sec:gld_description}

\begin{figure}[t]
    \centering
    \includegraphics[width=0.7\linewidth]{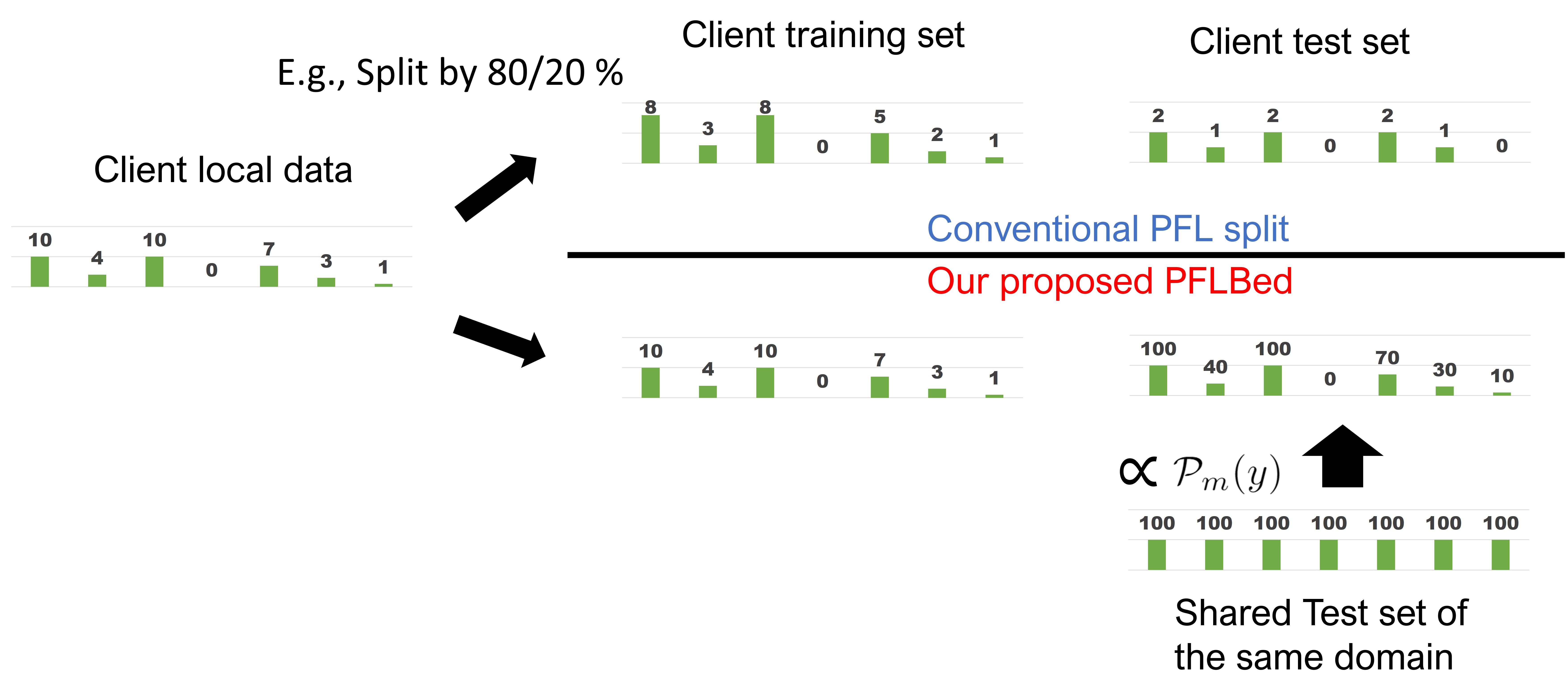}
    \caption{Illustration of the difference between traditional PFL split and our proposed \PFLBed in~\cref{s_pflbed} for a client. For the conventional way, given that each client may have limited data per class, after a training/test split, the distribution might be no longer matched, leading to an unfaithful evaluation. On the contrary, Our proposed way uses a shared test set from the same domain and re-weight the examples in evaluation by classes (\eg, weighted accuracy).}
    \label{fig:split}
\end{figure}

In~\cref{s_pflbed}, we provide several aspects including cross-domain and class non-IID $\sP_m(\vx, y)$, sufficient test samples, matched training/test splits, and distributional robustness evaluated with the class-balanced accuracy. We propose a standardized process called \PFLBed to construct a faithful personalized dataset for PFL algorithm development. As examples, we propose to transform some existing datasets including PACS and Office-Home, that are widely used in bench-marking domain adaption tasks, into PFL datasets. These datasets are suitable for experimental use in research since they are created with clear domain differences such as image styles like \emph{Photo} or \emph{Art}. The illustration~\cref{fig:split} shows the difference in preparing the test split for each client between the conventional PFL way and our proposed procedure based on \PFLBed. For the conventional way, given that each client may have limited data per class, after a training/test split, the distribution might be no longer matched, leading to an unfaithful evaluation. On the contrary, Our proposed way uses a shared test set from the same domain and re-weight the examples in evaluation by classes (\eg, weighted accuracy). Currently, for the sake of simplicity, we consider each client comes from one domain so the test set can simply be all the test images from that domain. We note that it is straightforward to make each client from a mixture of domains. 

In our experiments, for completeness, we also include the naturally partitioned dataset GLD-v2~\cite{weyand2020google,hsu2020federated}, a dataset consisting of landmark photographs taken from various locations around the world by different photographers where each partition contains a photographer's photos. We can view the style difference among the photographers as the domain gap thus treating each client as a domain.

\subsection{Visualizations of \PFLBed dataset client distribution}
\label{suppl-sec:dataset_description}
\begin{figure}[t]
    \centering
    \includegraphics[width=0.5\linewidth]{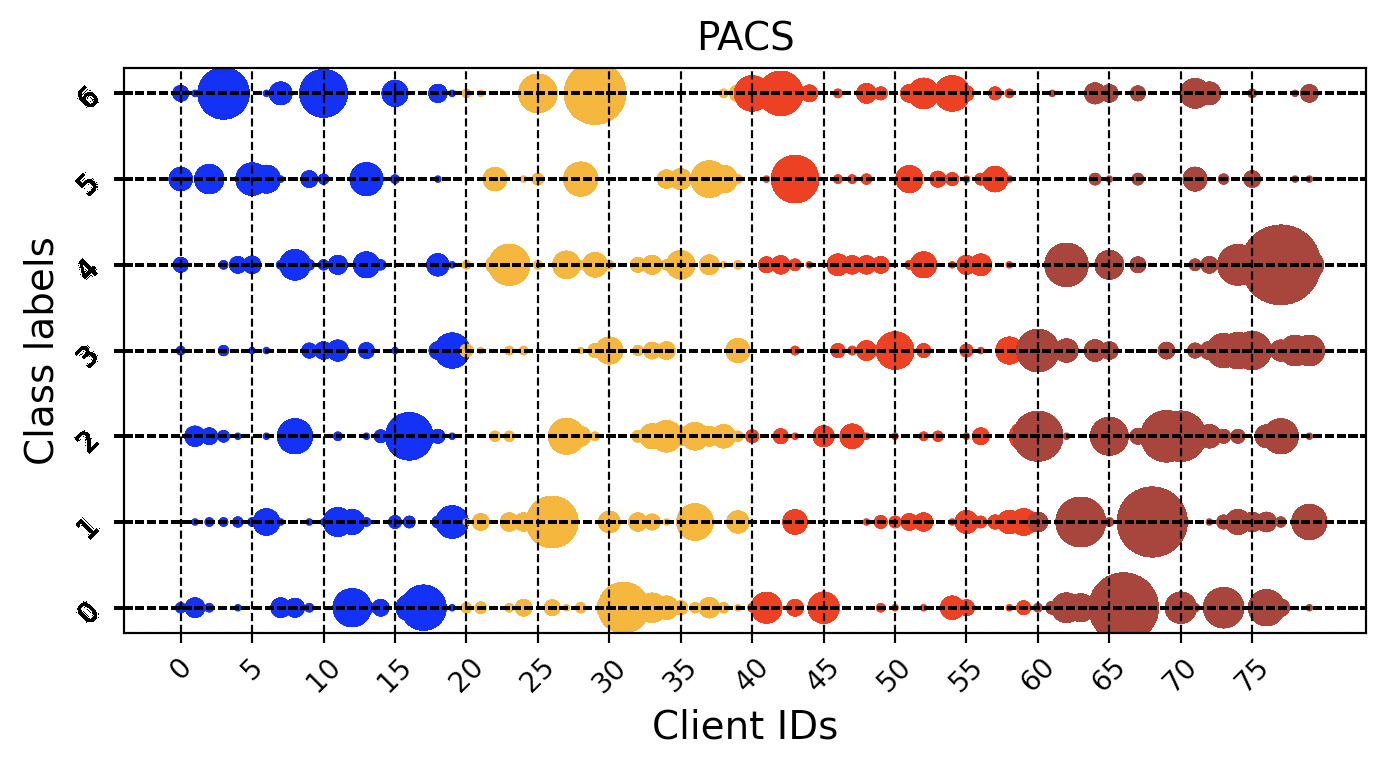}
    \vskip-5pt
    \caption{Clients distribution of PACS dataset across $4$ different domains. Each number on the horizontal axis represents a particular client for a total of $M=80$ clients. Each number on the vertical axis represents a particular class label for a total of $7$ classes. The maximum number of samples per class is $256$. }
    \label{fig:pacs_distribution}
\end{figure}

\begin{figure}[t]
    \centering
    \includegraphics[width=0.5\linewidth]{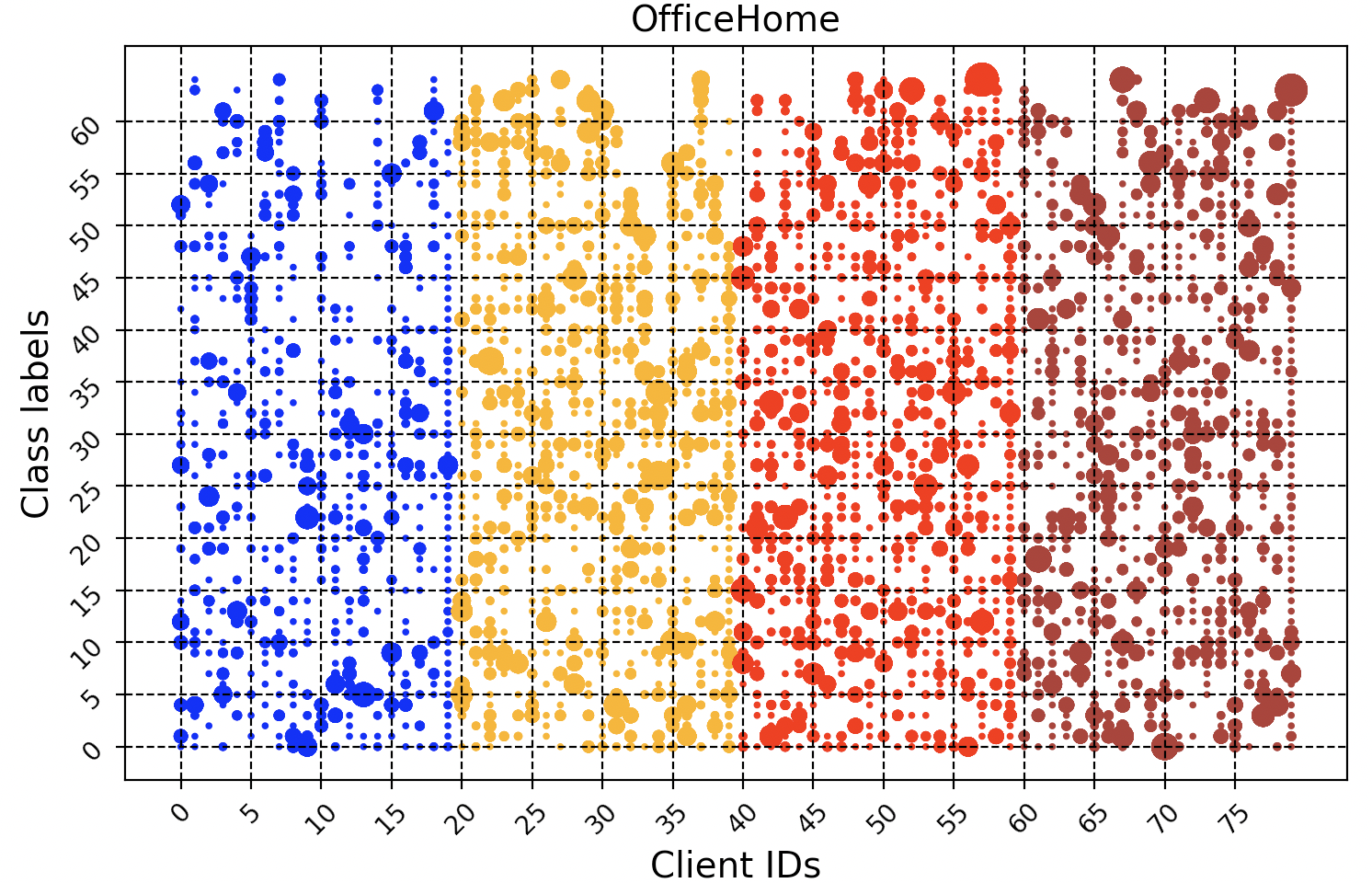}
    \caption{Clients distribution of Office-Home dataset across $4$ different domains. Each number on the horizontal axis represents a particular client for a total of $M=80$ clients. Each number on the vertical axis represents a particular class label for a total of $65$ classes. The maximum number of samples per class is $49$. }
    \label{fig:officehome_distribution}
\end{figure}

Here we show example client distributions of our proposed datasets for \PFLBed. For PACS and Office-Home datasets, we follow the procedures outlined in \cref{s_pflbed} where each client is sampled from Dirichlet($0.3$) within each domain. We visualize the distributions in~\cref{fig:pacs_distribution} and~\cref{fig:officehome_distribution} that the size of each point is proportional to the counts per class in a domain. Each column can be viewed as a single client's label distribution. As we can see, our clients show both label space $\sP_m(y)$ and domain space $\sP_m(x)$ heterogeneity. In~\cref{suppl-sec:gld_description} it shows the class distribution of the GLD dataset but it does not show domain differences through color differences since it is naturally non-IID without a specific domain annotation; each client can be directly treated as an independent domain.

\begin{figure}[t]
    \centering
    \includegraphics[width=0.5\linewidth]{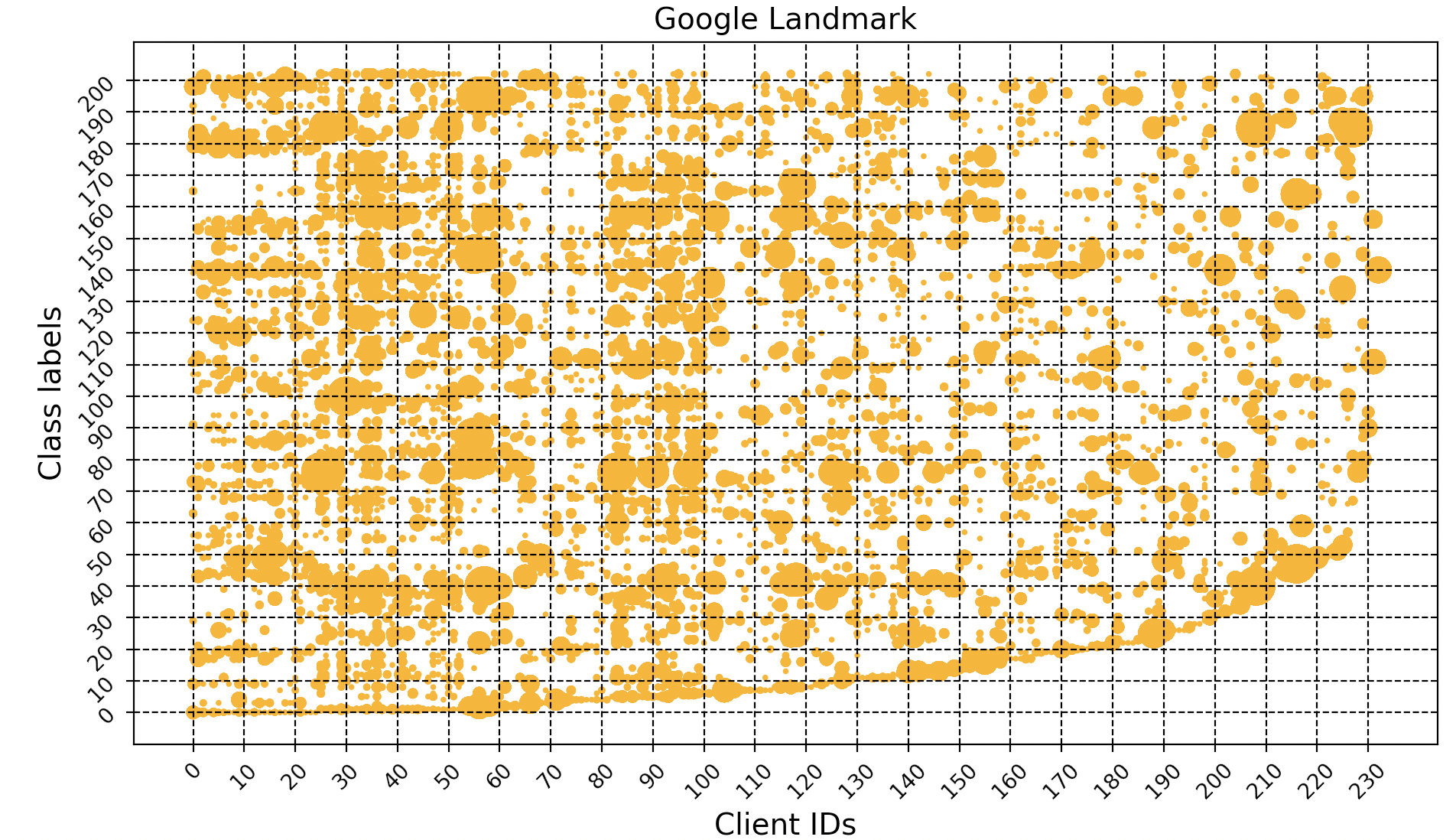}
    \caption{Clients distribution of GLD23k dataset. Each number on the horizontal axis represents a particular client for a total of $M=233$ clients. Each number on the vertical axis represents a particular class label for a total of $D=203$ classes. The maximum number of samples per class is 100. }
    \label{fig:gld_distribution}
\end{figure}

\clearpage
\section{More results and discussion}
\label{suppl-sec:results}

\subsection{Additional details and analyses of the bases collapse problem in~\cref{ss_algorithm}}
\label{suppl-ss-d1}

\begin{table}[t] 
    \caption{\small Baseline federated training algorithm for \FedBasis on PACS datasets with different \textbf{communication frequencies}. We consider $M=40$ participating clients and $K=4$ bases of ResNet-18 from random initialization (different from~\cref{s_exp} that use a pre-trained one that each basis model starts from the same initialization). To evaluate the model's personalized performance and general quality, we report personalized accuracy and global accuracy defined in~\cref{eq:acc}; both are averaged over clients.}
    \label{tab:Q3}
    \centering
    \footnotesize
    \setlength{\tabcolsep}{3pt}
	\renewcommand{\arraystretch}{1}
    \begin{tabular}{l|cc|cc}
    \toprule
    Method  & \multicolumn{2}{c|}{every iteration}  & \multicolumn{2}{c}{every $5$ epochs} \\
    \midrule
    Evaluation & Personalized & Global & Personalized & Global \\
    \midrule
    Test Accuracy &  90.2 & 66.0 & 69.0	& 68.8\\
    \bottomrule
    \end{tabular}
\end{table}

In~\cref{ss_algorithm} and \cref{fig:Q3} of the main paper, we presented a PFL experimental result to showcase the bases collapse problem of the baseline training algorithm (cf.~\cref{eq_baseline_local} and~\cref{eq_baseline_glob}). Here, we provide additional details and analyses that we omitted in the main paper due to the page limit.

\noindent\textbf{Brief experimental setup.}  We use the PACS~\cite{PACSli2017deeper} dataset, which contains in total $7$K training images from $7$ classes. We follow the procedure detailed in \cref{s_pflbed} to split the training images into $M=40$ non-IID clients. Each client has images from one of the four domains (Photo, Art, Cartoon, Sketch); the class distribution $\sP_m(y)$ of each client $m$ is sampled from a Dirichlet($0.3$) distribution to make it skewed and not identical among clients~\cite{hsu2019measuring}. We use $K=4$ bases and model each by a ResNet-18~\cite{he2016deep}. We apply the block-wise combinations introduced in~\cref{ss-extensino} to increase the representation power of the bases: different blocks are expected to capture different relationships of the image domains and class distributions jointly.
Different from the main studies in~\cref{s_exp}, to better measure the bases collapse problem, each basis model is randomly initialized rather than starting from the same pre-trained model. We train with the number of rounds equal to $100$ epochs overall, using the same training setups described in~\cref{suppl-sec:hyper}. 

\noindent\textbf{Evaluation.}
We consider two cases in which a model could perform poorly in a PFL setting: \textbf{1)} it suffers over-fitting, or \textbf{2)} it suffers under-fitting; \ie, not well-personalized to each client's data distribution.
We adopt two evaluation metrics to better contextualize the quality of the trained model. First, we follow the \PFLBed procedure in~\cref{s_pflbed} to prepare for each domain a ``class-balanced'' test set and re-weight it with $\sP_m(y)$ to calculate the personalized accuracy for client $m$. (As a reminder, each client has data from one single domain.) 
Second, we disregard the re-weighting step but directly evaluate each personalized model using the ``class-balanced'' test set of its corresponding domain. 
(That is, we directly calculate the performance on the global test set assigned in step 2 of the \PFLBed procedure in~\cref{s_pflbed} for each client model, without any re-weighting.) 

Without loss of generality, let us assume that each test set has the same number of test images $N_m$, and each test sample is indexed by $j$. 
The two metrics mentioned above can be formulated as:
\begin{align}
\label{eq:acc}
& \textbf{Personalized}\text{ accuracy:} && \frac{1}{M}\sum_m\frac{\sum_j\sP_m(y_j)\textbf{1}[y_j = h_{\vtheta_m}(\vx_j)]}{\sum_j\sP_m(y_j)},\nonumber
\\
& \textbf{Global}\text{ accuracy:} && \frac{1}{M}\sum_m\frac{1}{N_m}\sum_{j}{\textbf{1}[y_j = h_{\vtheta_m}(\vx_j)]}.
\end{align} 
The personalized accuracy weighs each test sample by $\sP_m(y_j)$ to reflect the class distribution of client $m$'s training data. This can be considered the standard personalized accuracy in literature. The global accuracy, in contrast, treats each test sample of client $m$'s domain equally. To summarize the accuracies of clients, we simply take the average over their accuracy.

Next, we consider the baseline training algorithm in~\cref{eq_baseline_local} and~\cref{eq_baseline_glob} for training our \FedBasis architecture with different communication frequencies.  

\noindent\textbf{Unlimited communication.} In terms of the number of local gradient steps per round and the number of total rounds (fixed to $100$ epochs of updates), we first consider an \emph{ideal} case: unlimited communication. This allows us to perform global aggregation as soon as we can; \ie, after each mini-batch SGD step. This training procedure very much recovers the conventional centralized training.

\noindent\textbf{Limited communication.} In practice, due to communication constraints, it is infeasible to perform global aggregation after each mini-batch SGD step. The standard FL setting is constrained by communication frequency due to the network transmission overload; clients typically can only communicate once after epochs of local SGD steps. We thus study the standard case~\cite{mcmahan2017communication}, performing local training for a few ($5$ here) epochs per round. \cref{tab:Q3} summarizes the results. We have the following observations. 

\begin{itemize} [topsep=1pt] 
    \item With unlimited communication (the ``every iteration'' column), \FedBasis achieves strong personalized accuracy, much higher than the global accuracy. 
    In other words, under the ideal case, we justify 1) the capacity of our convex combination representation and 2) the capability of the baseline training algorithm for producing personalized models dedicated to each clients' individual distributions.
    \item With limited communication (the ``every $5$ epochs'' column), \FedBasis in the more standard FL setting can no longer match the personalized accuracy in the unlimited communication setting.
    \item To our surprise, under the limited communication scenario, the personalized accuracy is comparable to the global accuracy. Namely, the constructed personalized models do learn good general knowledge (thus not over-fitting) but fail to personalize since they seem almost identical among clients.
\end{itemize}

These observations motivate our study and analyses in~\cref{ss_algorithm}. As confirmed in~\cref{fig:Q3}, \FedBasis by baseline training collapses to non-specialized bases and uniform combinations. In~\cref{fig:Q3}, we found that both the pairwise similarity and the entropy increase along with local training iterations and training rounds. We thus resolve the bases collapse issue by the proposed improved training algorithm in~\cref{eq_improved}. 

\subsection{Other baselines: Principal Component Analysis (PCA) and $k$-means clustering} Our \FedBasis architecture is to represent personalized models by a set of few basis models. Here, we present another baseline, building upon a reverse way of thinking: \emph{How can we summarize many personalized models into combinations of a few basis models given the federated constraint that no data are available at the server?} A straightforward way to achieve such model compression is to perform 
Principal Component Analysis (PCA) on the collection of all the personalized models. 
That is, we can represent each personalized model by the top-$k$ principal components (as $\{\vv_1, \dots, \vv_k\}$) found by PCA. 

We follow the experimental setup in~\cref{suppl-ss-d1} to construct $40$, non-IID clients. We consider an \emph{ideal} case of personalization in the unlimited communication setting. We first train a global model with mini-batches SGD, and then fine-tune it on each client's dataset to obtain $40$ personalized models $\{\vtheta_m\}$. Then, we perform PCA on their vectorized parameters. 

\begin{figure}[t]
    \centering
    \includegraphics[width=0.25\linewidth]{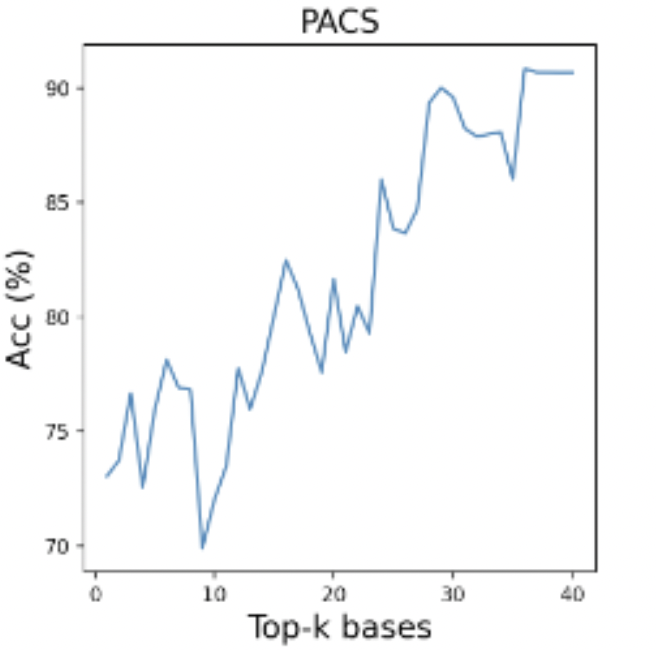}
    \vskip-5pt
    \caption{Reducing $40$ fine-tuned personalized models into top-$k$ bases by PCA.}
    \label{fig:pca}
    \vskip-10pt
\end{figure}

As shown in~\cref{fig:pca}, we observe that the averaged personalized performance drops drastically as the number of eigenvectors decreases. For instance, using only the top-$4$ bases leads to slumps in the accuracy of $18.1\%$ for PACS. It demonstrates the challenge of this problem. We hypothesize that the poor performance is likely due to (1) personalized models produced by fine-tuning do not simply lie on a low-dimensional subspace and/or (2) PCA in the model parameters cannot guarantee that the reconstructed models maintain their accuracies. More specifically, PCA aims to minimize the difference between the original models and the reconstructed models in their model parameters, not their accuracies on the personalized test data. 
As a result, we can observe some fluctuations in accuracy along with the changes in the value $k$.  

Alternately, we investigate using $k$-means clustering on the personalized models $\{\vtheta_m\}$ parameters to cluster them into $k=4$ models and use each client's assigned centroid as the personalized models. We again see a significant accuracy drop of $21.4\%$ for PACS. 

Therefore, we are motivated to solve our proposed objective~\cref{eq:fedbasis} that aims to directly learn the bases such that all personalized models can be their linear combinations while minimizing the local empirical risks. 

\subsection{More results about the robustness of \FedBasis}
\begin{figure}[t]
    \centering
    \minipage{0.5\linewidth}
    \centering
    \includegraphics[width=0.7\linewidth, height=4.5cm]{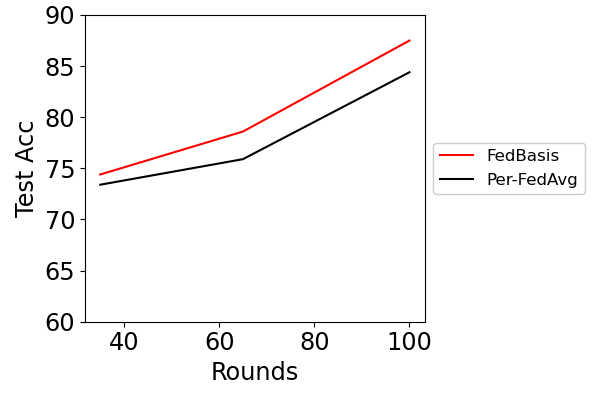}
    \vfill
    \mbox{\small (a)}
    \endminipage
    \hfill
    \minipage{0.5\linewidth}
    \centering
    \includegraphics[width=0.8\linewidth, height=4.5cm]{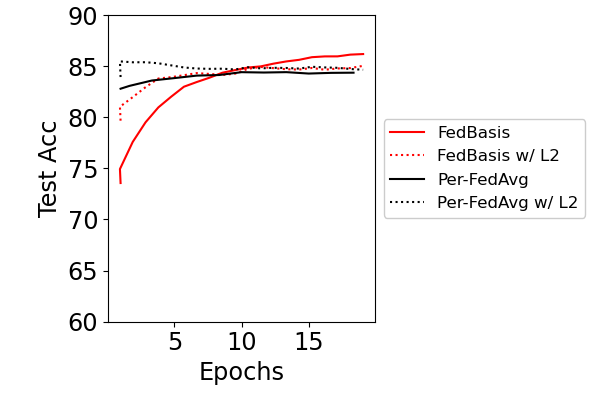} 
    \vfill
    \mbox{\small (b) }
    \endminipage
    \caption{(a) Federated training curves of Office-Home (small) dataset (\cref{tab:noniid} in the main paper) of \PerFedAvgFT and \FedBasis along the rounds. In each evaluated round (note that \FedBasis warm-starts from $30$ rounds of \FedAvg, as described in~\cref{ss-extensino}), we run the adaption procedure to evaluate the new clients to report the averaged personalized accuracy as the same as~\cref{tab:noniid}. (b) Fine-tuning curves of Office-Home (small) dataset (\cref{tab:noniid} in the main paper) of \PerFedAvgFT and \FedBasis with fine-tuning learning rate $=0.005$ with the $\ell$-2 regularization in Equation 2 of~\cite{dinh2020personalized} (see~\cref{tab:reg}).} 
    \label{fig:oh_l2}
\end{figure}

In both~\cref{tab:noniid} in the main paper, we demonstrate the robustness of the \FedBasis on the choices of stopping epochs when it is fine-tuned for new clients, compared to other baselines. Note that, in the current~\cref{tab:noniid}, for each method and each dataset, we highlight the difference ($|\Delta|$) between stopping the fine-tuning by the last epoch or by the best epoch selected by validation. In~\cref{fig:vis}, we further compare the best baseline \PerFedAvgFT with different learning rates. We observe \FedBasis is clearly more robust to hyperparameters such as the learning rates and stopping epoch.  We focus on the more challenging datasets Office-Home with the small training size setting. In~\cref{fig:oh_l2}, we plotted out the dynamics of the federated training and regularized fine-tuning on new clients, both demonstrating the effectiveness of our \FedBasis. We attribute it to the clear advantage that \FedBasis only needs to personalize much fewer parameters when adapting to a new client, thus enjoying the robustness. We further note for \PerFedAvgFT, although with proper tuning it can achieve decent performance (still lower than ours), this requires a validation set for each client thus likely not practical in the real world.

\subsection{Extended comparisons and results of~\cref{tab:noniid}}

\paragraph{Variants of \PerFedAvg.} In~\cref{s_exp}, we focus on the first-order (\textbf{FO}) version of \PerFedAvgFT~\cite{fallah2020personalized} due to its better accuracy and training efficiency on the datasets in our experiments. In~\cref{tab:pfedavvg}, we provide a comparison on \PerFedAvgFT with the two variants FO and Hessian-free (HF) introduced in~\cite{fallah2020personalized} and we confirmed FO is better in the performance.

\paragraph{Further fine-tuning as personalization components.}
In~\cref{s_exp}, we consider personalization for new clients using the proposed way of each method. Here we provide more complete results (due to the limited space in the main paper) by further considering fine-tuning the classifiers (as linear probe) or the whole model (full fine-tuning, FL in short). As shown in~\cref{sup-tab:noniid_more}, overall \FedBasis still performs the most competitively in terms of accuracy and robustness. Overall, although linear probes are sometimes more robust than full fine-tuning,  full fine-tuning can typically outperform linear probes with proper validation. \FedBasis somewhat provide a nice balance for such a dilemma since it already learns to personalize by basis composition in the training phase, thanks to the expressive power of non-linear deep neural networks. 

\paragraph{Random seeds variances of~\cref{tab:noniid}.}
We provide the variances of 3 different runs with different random seeds in~\cref{sup-tab:noniid_var} due to the limited space in the main paper.

\paragraph{Effects of numbers of bases $K$.} We study the effects of numbers of bases $K$. We follow the experiments in~\cref{tab:noniid} to use the small local size and select the stopping epoch by validations. As shown in~\cref{sup-tab:bases}, a small number of bases is enough to accommodate the personalized variation among the clients, thanks to the effective training in \FedBasis that enforce the bases to learn to be expressive for combining into different personalized models.  

\paragraph{Ablations.} We show in~\cref{ss-tab:abl} that our major basis introduced in~\cref{ss-extensino} improves the performance, along with other designs we proposed in~\cref{ss_algorithm}.

\begin{table}[t]
    \caption{More results of \PerFedAvgFT with first-order (FO) and Hessian-free (HF) variants in~\cite{fallah2020personalized}. The results on based on Office-Home (\textbf{M}oderate training size).}
    \label{tab:pfedavvg}
    \centering
    \footnotesize
    \setlength{\tabcolsep}{2.5pt}
	\renewcommand{\arraystretch}{1}
    \begin{tabular}{c|cccccc}
    \toprule
    Method & \multicolumn{5}{c}{Last/Best Acc.}\\
    Learning rates & $0.001$ & $0.005$ & $0.01$ & $0.05$ & $0.1$\\
    \midrule
    \FedAvgFT & 78.1/78.1 &78.2/78.5& 70.5/76.6 &65.4/75.6 & 38.1/73.1\\
    \PerFedAvgFT (FO) & 86.1/86.1 &86.1/86.2 & 65.3/85.4& 60.5/83.4 & 40.1/82.6\\
    \PerFedAvgFT (HF) & 85.3/85.3 &85.5/85.5 & 63.4/83.6 & 43.7/81.6 & 34.8/80.7 \\
    \FedBasis & 87.6/87.6 &87.5/87.7 & 87.6/87.7 & 87.6/87.6 & 87.5/87.6\\
    \bottomrule
    \end{tabular} 
\end{table}

\begin{table*}[t]
    \caption{\small More comprehensive results of~\cref{tab:noniid} with further personalization. \textbf{LP:} further training for a personalized linear classifier. \textbf{FT:} further fine-tuning for the whole model. Averaged personalized test accuracy ($\%$) on class non-IID new clients sampled from Dirichlet(0.3). Each method is learned on each client's training data of different sizes for $20$ epochs with a learning rate selected from $\{0.005, 0.01, 0.05\}$. We report both the \textbf{Last} epoch and the \textbf{Best} by validation. }
 
    \label{sup-tab:noniid_more}
    \centering
    \footnotesize
    \setlength{\tabcolsep}{2.5pt}
	\renewcommand{\arraystretch}{0.75}
    \begin{tabular}{l|cc|a|cc|a|cc|a|cc|a|cc|a|cc|a|cc|a}
    \toprule 
    Method/Dataset  & \multicolumn{6}{c}{PACS} & \multicolumn{6}{c}{Office-Home} & \multicolumn{9}{c}{GLD23k}\\
    \midrule
    Training Size  & \multicolumn{3}{c}{\textbf{S}} & \multicolumn{3}{c|}{\textbf{M}} & \multicolumn{3}{c}{\textbf{S}} & \multicolumn{3}{c|}{\textbf{M}} & \multicolumn{3}{c}{\textbf{S}} & \multicolumn{3}{c}{\textbf{M}} & \multicolumn{3}{c}{\textbf{L}}\\
    \midrule
    Epoch  & Last & Best & $|\Delta|$ & Last & Best &$|\Delta|$ & Last & Best &$|\Delta|$ & Last & Best &$|\Delta|$ & Last & Best &$|\Delta|$ & Last & Best &$|\Delta|$ & Last & Best &$|\Delta|$\\
    \midrule
    \FedRepLP  & 87.4 & 87.4 & 0.0 & 92.5 & 92.4 & 0.1 & 75.6 & 75.6 & 0.0 & 76.0 & 76.1 & 0.1 &75.7 & 77.6 & 1.9 & 78.8 & 78.8 & 0.0 & 80.1 & 80.8 & 0.7\\
    \FedRepFT  & 89.8 & 89.8 & 0.0 & 92.4 & 92.5 & 0.1 & 74.2 & 76.1 & 1.9 & 75.2 & 76.4 & 1.2& 79.2 & 79.9 & 0.7& 81.5 & 82.5 & 1.0& 81.5 & 83.5 & 2.0\\
    \FedBNLP  & 86.2 & 88.2 & 2.0 & 92.4 & 92.4 & 0.0 & 76.9 & 77.0 & 0.1 & 78.1 & 78.1 & 0.0 & 74.1 & 74.5 & 0.4 & 76.6 & 76.6 & 0.0 & 76.4 & 76.5 & 0.1\\
    \FedBNFT  & 90.8 & 92.1 & 1.3 & 93.0 & 93.1 & 0.1 & 82.3 & 82.5 & 0.2 & 79.0 & 79.2 & 0.2 & 68.1 & 70.5 & 2.4 & 77.8 & 81.8 & 4.0 & 80.5 & 83.9 & 3.4 \\
    \midrule
    \pFedHN &  \multicolumn{2}{c|}{85.4} & - & \multicolumn{2}{c|}{85.5} & - & \multicolumn{2}{c|}{74.1} & - & \multicolumn{2}{c|}{74.3} & - & \multicolumn{2}{c|}{74.5} & - & \multicolumn{2}{c|}{75.6} & - & \multicolumn{2}{c|}{77.2} & -\\
    \pFedHNLP  & 90.4 & 90.4 & 0.0 & 90.6 & 90.6 & 0.0 & 75.1 & 75.1 & 0.0 & 77.4 & 77.4 & 0.0 & 77.0 & 77.6 & 0.6 & 78.5 & 78.5 & 0.0 & 79.1 & 79.5 & 0.4\\
    \pFedHNFT & 90.5 & 91.2 & 0.7 & 90.4 & 91.4 & 1.0 & 76.2 & 77.2 & 1.0 & 77.1 & 77.6 & 0.5 & 77.6 & 81.4 & 3.8 & 78.6 & 81.6 & 3.0 & 80.2 & 82.2 & 2.0\\    
    \PerFedAvgFT  & \textbf{95.4} & \textbf{95.6} & 0.2 & \textbf{96.2} & \textbf{96.3} & 0.1 & 84.3 & 84.4 & 0.1 & 86.1 & 86.2 & 0.1 & 78.5 & 85.3 & 6.8 & 79.9 & 85.2 & 5.3 & 82.2 & 86.1 & 3.9\\
    \midrule   
    \kNNPer  & \multicolumn{2}{c|}{71.6} & - & \multicolumn{2}{c|}{71.6} & - &  \multicolumn{2}{c|}{50.4} & - & \multicolumn{2}{c|}{54.5} & - & \multicolumn{2}{c|}{54.0} & - & \multicolumn{2}{c|}{57.4} & - & \multicolumn{2}{c|}{69.2} & -\\
    \kNNPerFT & 72.7 & 72.7 & 0.0 & 79.4 & 79.7 & 0.3 & 51.6 & 52.4 & 0.8 & 54.2 & 54.4 & 0.2 & 54.2 &54.5 & 0.3 & 57.1 & 57.8 & 0.7 & 69.5 & 70.2 & 0.7\\ 
    \midrule  
    \FedAvg & \multicolumn{2}{c|}{88.1} & - & \multicolumn{2}{c|}{88.1} & - & \multicolumn{2}{c|}{73.1} & - & \multicolumn{2}{c|}{73.1} & - & \multicolumn{2}{c|}{45.4} & - & \multicolumn{2}{c|}{45.4} & - & \multicolumn{2}{c|}{45.4} & -\\
    \FedAvgLP  & 88.2 & 90.1 & 1.9 & 90.5 & 90.5 & 0.0 & 76.6 & 76.6 & 0.0 & 77.0 & 77.0 & 0.0 & 80.8 & 81.5 & 0.7 & 80.9 & 81.8 & 0.9 & 83.3 & 83.3 & 0.0\\
    \FedAvgFT  & 86.1 & 91.9 & 5.8 & 90.5 & 90.5 & 0.0 & 76.1 & 77.4 & 1.3 & 78.2 & 78.5 & 0.3 & 81.5 & 84.2 & 2.7 & 81.6 & 84.5 & 2.9 & 84.1 & 86.1 & 2.0\\
    \midrule     
    \FedBasis & 95.2 & 95.2 & 0.0 & \textbf{96.2} & 96.2 & 0.0 & \textbf{87.4} & \textbf{87.5} & 0.1 & \textbf{87.5} & \textbf{87.7} & 0.2 & \textbf{87.4} & \textbf{87.4} & 0.0 & \textbf{87.6} & \textbf{87.6} & 0.0 & \textbf{89.0} & \textbf{89.1} & 0.1\\

    \bottomrule
    \end{tabular}
\end{table*}

\begin{table}[h] 
    \caption{\small Variance ($\sigma^{2}$) of personalized test accuracy ($\%$) over $3$ different runs for~\cref{sup-tab:noniid_more}.}
    \label{sup-tab:noniid_var}
    \centering
    \footnotesize
    \setlength{\tabcolsep}{2.5pt}
	\renewcommand{\arraystretch}{0.75}
    \begin{tabular}{l|cccc|cccc|cccccc}
    \toprule 
    Method/Dataset  & \multicolumn{4}{c}{PACS} & \multicolumn{4}{c}{Office-Home} & \multicolumn{6}{c}{GLD23k}\\
    \midrule
    Training Size  & \multicolumn{2}{c}{\textbf{S}} & \multicolumn{2}{c|}{\textbf{M}} & \multicolumn{2}{c}{\textbf{S}} & \multicolumn{2}{c|}{\textbf{M}} & \multicolumn{2}{c}{\textbf{S}} & \multicolumn{2}{c}{\textbf{M}} & \multicolumn{2}{c}{\textbf{L}}\\
    \midrule
    Epoch  & Last & Best & Last & Best & Last & Best & Last & Best & Last & Best & Last & Best & Last & Best\\
    \midrule
    \FedRepLP  & 0.22 & 0.16 & 0.21 & 0.22 & 0.31 & 0.26 & 0.33 & 0.25  & 0.55 & 0.56 & 0.47 & 0.52 & 0.44 & 0.29\\
    \FedRepFT  & 0.36 & 0.29 & 0.41 & 0.38 & 0.66 & 0.56 & 0.71 & 0.39 & 0.78 & 0.89 & 0.88 & 0.75 & 0.74 & 0.71\\
    \FedBNLP  & 0.15 & 0.16 & 0.31 & 0.15 & 0.12 & 0.23 & 0.28 & 0.19 & 0.36 & 0.41 & 0.29 & 0.21 & 0.51 & 0.39\\
    \FedBNFT  & 0.33 & 0.45 & 0.41 & 0.42 & 0.67 & 0.59 & 0.55 & 0.62 & 0.68 & 0.66 & 0.56 & 0.48 & 0.65 & 0.62\\
    \midrule
    \pFedHN & 0.78 & 0.64 & 0.56 & 0.57 & 0.46 & 0.51 & 0.48 & 0.55 & 0.41 & 0.28 & 0.55 & 0.56 & 0.39 & 0.28\\
    \pFedHNLP  & 0.36 & 0.44 & 0.29 & 0.36 & 0.27 & 0.31 & 0.28 & 0.25 & 0.87 & 0.86 & 0.82 & 0.75 & 0.78 & 0.58\\
    \pFedHNFT & 0.85 & 0.97 & 0.56 & 0.77 & 0.77 & 0.75 & 0.64 & 0.70 & 1.01 & 1.12 & 0.89 & 0.88 & 0.91 & 1.15\\    
    \PerFedAvgFT  & 0.51 & 0.46 & 0.37 & 0.41 & 0.70 & 0.61 & 0.63 & 0.66 & 0.51 & 0.25 & 0.48 & 0.45 & 0.34 & 0.38\\
    \midrule   
    \kNNPer  & 0.30 & 0.34 & 0.19 & 0.38 & 0.56 & 0.58 & 0.52 & 0.57 & 1.14 & 1.56 & 1.28 & 0.85 & 0.95 & 0.95\\
    \kNNPerFT & 1.75 & 1.41 & 1.25 & 1.39 & 0.57 & 0.60 & 0.78 & 0.69 & 0.27 & 0.56 & 0.48 & 0.71 & 0.55 & 0.61\\ 
    \midrule  
    \FedAvg & 0.23 & 0.25 & 0.29 & 0.24 & 0.38 & 0.41 & 0.50 & 0.42 & 0.54 & 0.39 & 0.55 & 0.56 & 0.54 & 0.27\\
    \FedAvgLP  & 0.15 & 0.21 & 0.20 & 0.17 & 0.29 & 0.22 & 0.31 & 0.44 & 0.63 & 0.65 & 0.48 & 0.59 & 0.64 & 0.59\\
    \FedAvgFT  & 0.52 & 0.39 & 0.44 & 0.51 & 0.57 & 0.46 & 0.60 & 0.58 & 0.68 & 0.71 & 0.59 & 0.58 & 0.61 & 0.48\\
    \midrule     
    \FedBasis & 0.52 & 0.56 & 0.45 & 0.50 & 0.38 & 0.39 & 0.42 & 0.45 & 0.52 & 0.66 & 0.47 & 0.68 & 0.29 & 0.45\\

    \bottomrule
    \end{tabular}
\end{table}

\begin{table}
    \caption{\small Effects of number of bases $K$ in \FedBasis. We use the small local size and select the stopping epoch by validations, according to~\cref{tab:noniid}. }
    \label{sup-tab:bases}
    \centering
    \footnotesize 
    \setlength{\tabcolsep}{1.75pt}
	\renewcommand{\arraystretch}{1}
    \begin{tabular}{c|cc}
    \toprule
     $K$ & PACS & Office-Home \\
    \midrule
    1 & 91.9 & 77.4 \\
    2 & 93.4 & 84.5 \\
    4 & 95.2 & 87.5 \\
    6 & 94.7 & 87.6 \\
    8 & 95.0 & 87.4 \\
    \bottomrule
    \end{tabular}
    \vskip -10pt
\end{table}

\begin{table}
    \caption{\small Ablation studies of \FedBasis designs in~\cref{s_approach}.}
    \label{ss-tab:abl}
    \centering
    \footnotesize 
    \setlength{\tabcolsep}{1.5pt}
	\renewcommand{\arraystretch}{0.75}
    \begin{tabular}{ccc|cc}
    \toprule
    \texttt{CoordinateDescent} & \texttt{MajorBasis} & $\tau$ & Office-Home & GLD \\
    \midrule
    \xmark & \cmark & 0.1 & 83.5 & 85.8 \\
    \cmark & \xmark & 0.1 & 87.2 & 83.3 \\
    \cmark & \cmark & 1.0 & 87.1 & 85.5 \\
    \cmark & \cmark & 0.1 & 87.5 & 87.6 \\
    \bottomrule
    \end{tabular}
    \vskip -10pt
\end{table}

\end{document}